%% file: main.tex
\documentclass[preprint,12pt]{elsarticle}
\usepackage[top=1in,bottom=1in,left=1in,right=1in]{geometry}
\usepackage{amssymb}
\usepackage{amsmath}
\usepackage{mathtools}
\usepackage{xcolor}
\usepackage{subfigure}
\usepackage{hyperref}
\usepackage{float} 

\usepackage{bm}   

\usepackage{adjustbox} 
\usepackage{multirow} 
\usepackage{subcaption} 

\biboptions{sort&compress}

\begin{document}

\begin{frontmatter}

\title{A comprehensive and FAIR comparison between MLP and KAN representations for differential equations and operator networks}


\author[brown]{Khemraj Shukla\fnref{1}}
\author[brown]{Juan Diego Toscano\fnref{1}}
\author[brown]{Zhicheng Wang\fnref{1}}
\author[brown]{Zongren Zou\fnref{1}}
\author[brown,PNNL]{George Em Karniadakis \fnref{2}}

\fntext[1]{These authors contributed equally to this work and are listed in alphabetical order by last name.}
\fntext[2]{Corresponding author: george\_karniadakis@brown.edu (George Em Karniadakis).}
\address[brown]{Division of Applied Mathematics, Brown University, Providence, RI 02906, USA}
\address[PNNL]{Pacific Northwest National Laboratory, Richland, WA 99354, USA}

\begin{abstract}
Kolmogorov-Arnold Networks (KANs) were recently introduced as an alternative representation model to MLP. Herein, we 
employ KANs to construct physics-informed machine learning models (PIKANs) and deep operator models (DeepOKANs) for solving differential equations for forward and inverse problems. In particular, we compare them with physics-informed neural networks (PINNs) and deep operator networks (DeepONets), which are based on the standard MLP representation. We find that although the original KANs based on the B-splines parameterization lack accuracy and efficiency, modified versions based on low-order orthogonal polynomials have comparable performance to PINNs and DeepONet although they still lack robustness as they may diverge for different random seeds or higher order orthogonal polynomials. We visualize their corresponding loss landscapes and analyze their learning dynamics using information bottleneck theory. Our study follows the FAIR principles so that other researchers can use our benchmarks to further advance this emerging topic.



\end{abstract}


\end{frontmatter}

\section{Introduction}
Multilayer perceptrons (MLPs) are a class of feedforward artificial neural networks consisting of at least three layers of nodes: an input layer, one or more hidden layers, and an output layer \cite{haykin1998neural,cybenko1989approximation}. As stated in the universal approximation theorem \cite{hornik1989multilayer}, MLPs can learn non-linear relationships and patterns in data, making them one of the main building blocks of modern deep learning applications \cite{goodfellow2020generative,vaswani2017attention,he2016deep,li2018visualizing,toscano2023teeth,kaelbling1996reinforcement}. However, due to their complex and deeply nested structure, MLPs lack interpretability \cite{cranmer2023interpretable} and often face challenges such as overfitting, vanishing or exploding gradients, and scalability issues. 

As an alternative to MLP, researchers have recently proposed Kolmogorov-Arnold Networks (KANs), a new type of model that aims to be more accurate and interpretable than MLP \cite{liu2024kan}. KANs are inspired by the Kolmogorov-Arnold representation theorem and can be interpreted as a combination of Kolmogorov Networks \cite{sprecher2002space,koppen2002training,schmidhuber1997discovering,lai2021kolmogorov,leni2013kolmogorov,he2023optimal} and MLPs with learnable activation functions \cite{jagtap2020adaptive,guarnieri1999multilayer,fakhoury2022exsplinet}. KANs and their rapidly growing extensions have shown promising performance in addressing several MLP issues, such as interpretability and catastrophic forgetting in supervised and unsupervised learning tasks \cite{liu2024kan,vaca2024kolmogorov,samadi2024smooth}. However, their formulation relies on learnable B-Splines as activation functions, significantly increasing their computational cost. To address this problem, several subsequent studies proposed using alternative univariate functions such as radial basis functions (RBF)~\cite{li2024kolmogorov}, wavelets~\cite{bozorgasl2024wavkan}  or Jacobi polynomials \cite{jacobiKANs,chebykan,ss2024chebyshev,torchkan} (e.g., Chebyshev, Legendre).

KANs have also been explored for solving differential equations and operator learning. Liu et al. \cite{liu2024kan} combined physics-informed neural networks (PINNs) \cite{raissi2019physics} and KANs to solve a 2D Poison equation. Similarly, Abueida et al. proposed DeepOKAN \cite{abueidda2024deepokan}, an RFB-based KAN operator network, to solve a 2D orthotropic elasticity problem. The authors in \cite{liu2024kan} and \cite{abueidda2024deepokan} show that KANs significantly outperform MLP; however, these studies were limited to shallow networks and were conducted on simplified problems.

Herein, we employ physics-informed machine learning \cite{karniadakis2021physics} and KANs to develop PIKANs and operator models (DeepOKANs).
In the first part of this study, we systematically compare PINN and PIKAN variations on six benchmarks carefully selected from the current literature \cite{mcclenny2023self,wang2023solution, lu2021learning,wu2023comprehensive}. To allow a fair comparison between these representation models, we combine them with state-of-the-art optimization techniques such as residual-based attention \cite{anagnostopoulos2024learning} and eddy viscosity formulations \cite{wang2023solution}, which enable these new models to solve more complex problems. In this section, we analyze PIKAN stability and sensitivity to higher polynomial orders and the number of layers. In the second part, we compare DeepOKANs and DeepONets for two operator learning tasks. 

In the last section of this paper, we analyze PIKAN learning dynamics through the lens of the information bottleneck (IB) theory \cite{tishby2000information,tishby2015deep}. According to the IB theory, a well-functioning model should retain essential output information while discarding insignificant input details, thereby creating an ``information bottleneck'' that induces two distinct stages of training ``fitting'' and ``diffusion'' \cite{shwartz2017opening,goldfeld2020information}. Recently, Anagnostopoulos et al., \cite{anagnostopoulos2024learning} extended this theory to PINNs and proposed the existence of a third phase named ``total diffusion". Following \cite{anagnostopoulos2024learning}, we analyze the PIKAN training dynamics and identify the three stages of learning observed in PINNs, bridging the gap between both representation models.

This paper is organized as follows. In Section \ref{Sec2}, we briefly describe the problem and the representation models. Section \ref{Sec3} compares the performance of both representation models in eight benchmarks, including discontinuous function approximation, structure-preserving Hamiltonian dynamical systems, PDE solution approximation, uncertainty quantification, and operator learning. Finally, Section \ref{Sec6} analyzes the training dynamics of PINNs and PIKANs based on the IB method. We summarize in Section \ref{sec:summary}.

\input{sec_2}
\input{sec_3}

\input{sec_4}

\section{Summary}\label{sec:summary}
This study investigates the potential and effectiveness of KAN-based representations for tackling key challenges in scientific machine learning. We begin with  evaluating KAN and Chebyshev-KAN for approximating discontinuous and oscillatory functions. The investigation revealed that both MLPs and KANs achieved high accuracy in approximating the function. However, training KANs was significantly slower compared to MLP representations. Notably, the Chebyshev-KAN approach exhibited unstable training, leading to rapid divergence. To address this issue, we implemented a modification to the Chebyshev-KAN forward pass by composing it with $\tanh$ function. This modification successfully stabilized the training process. Consequently, the function approximation performance of the modified Chebyshev-KAN became comparable to MLPs in terms of both accuracy and runtime efficiency. We then explored the application of Chebyshev-KAN to problems involving structure-preserving and energy-conserving dynamical systems, comparing its efficacy to MLP-based architectures. The results showed that Chebyshev-KAN is not as data-efficient as MLPs. However, with a slightly larger amount of training data, the performance of both models became comparable. Next, we consider the 2D Helmholtz equation. In this example, we demonstrate the capabilities of various neural network architectures, including PINN, PIKAN, PIKAN (multigrid), cPIKAN, PIKAN with RBA, and cPIKAN with RBA to recover the highly oscialltory solutions. Additionally, we analyze the loss landscapes for all these architectures to investigate their convexity and convergence behavior towards minima across different landscapes. Next we consider 2D steady state incompressible Navier-Stokes equation to solve the lid-driven cavity flow problem for low to high Reynold's number. Here in addition to Chebyshev, we also use Jacobi \cite{karniadakis2005spectral} and Legendre polynomials \cite{karniadakis2005spectral}. The Jacobi polynomials based KANs are proved to be promising candidate for physics informed network predict the incompressible flow.  Among the various polynomials we have tested, the Chebyshev PIKAN (cPIKAN) shows its advantage in terms of accuracy and training time. Compared to the PINN, cPIKAN can achieve the same accuracy with far less network parameters, but it takes more GPU hours to train. The residual-based attention (RBA) helps to improve both PINN and PIKAN's accuracy, without need to consume more training time. In addition, the vanila PIKANs suffered from unstable training for flow at $Re$, shown by the fact that the relative error barely goes down with training. Nonetheless, with the help of the entropy viscosity method  (EVM), it can recover the correct training trajectory and obtain accurate solution.

Next we study the KAN based representation for solving the PDE with noisy data. In addressing noisy data with uncertainty quantification in solving differential equations, cPIKANs are compatible with the Bayesian framework and the Bayesian cPIKAN (B-cPIKAN) method is able to provide similar predicted mean and uncertainty as the Bayesian PINN (B-PINN) method \cite{YANG2021109913} at higher computational cost. However, specifying the prior distribution for parameters of the B-cPIKAN is not as straightforward as it is for parameters of the B-PINN, and requires further theoretical and numerical work in the future.
As discussed in Section \ref{Sec3}, the DeepOKAN, which integrates the DeepONet \cite{lu2021learning} structure with the Chebyshev KAN architecture \cite{chebykan}, has shown competitive performance in operator learning compared to the DeepONet, indicating DeepOKAN as a promising alternative representation model. Additionally, it is significantly more robust to noisy input functions in the testing stage after being trained with clean data. 

Finally, we study the learning behavior of KAN and MLP-based representation using information-bottleneck theory. The Information Bottleneck (IB) method has been effectively extended to the study of cPIKANs. The foundational training dynamics exhibit remarkable similarities despite the architectural distinctions between PINNs and cPIKANs. Both representation models demonstrate a consistent progression through the stages of fitting, diffusion, and total diffusion, as outlined by the IB framework. This insight into network behavior aims to bridge the gap between the representation models and motivate future research to use the IB method as a guide to develop training strategies and new architectures or enhance model performance.
Future research directions include
\begin{enumerate}
\item  Implementing KAN-based representations for solving large-scale partial differential equations (PDEs) using domain decomposition techniques. This approach has the potential to improve the scalability of KAN-based methods for complex problems.
\item  Implementing KAN-based representations for solving large-scale partial differential equations (PDEs) using domain decomposition techniques \cite{shukla2021parallel}. This approach has the potential to improve the scalability of KAN-based methods for complex problems.
\item Applying KAN-based representations to time-dependent PDEs in two and three dimensions. This would extend the applicability of KAN methods to a wider range of scientific and engineering problems.
\item A rigorous theory for the convergence of KAN for elliptic and parabolic class of PDEs.

\item Extending the applicability of DeepOKAN based architecture for formulating surrogate for industrial complexity problems \cite{shukla2024deep}.
\end{enumerate}

\bibliographystyle{elsarticle-num} 
\bibliography{references}

\end{document}

%% file: sec_2.tex
\section{Problem Formulation and representation models}
\label{Sec2}

\begin{figure}[h!]
    \centering
    \subfigure[Physics-informed networks.]{
    \includegraphics[scale=.27]{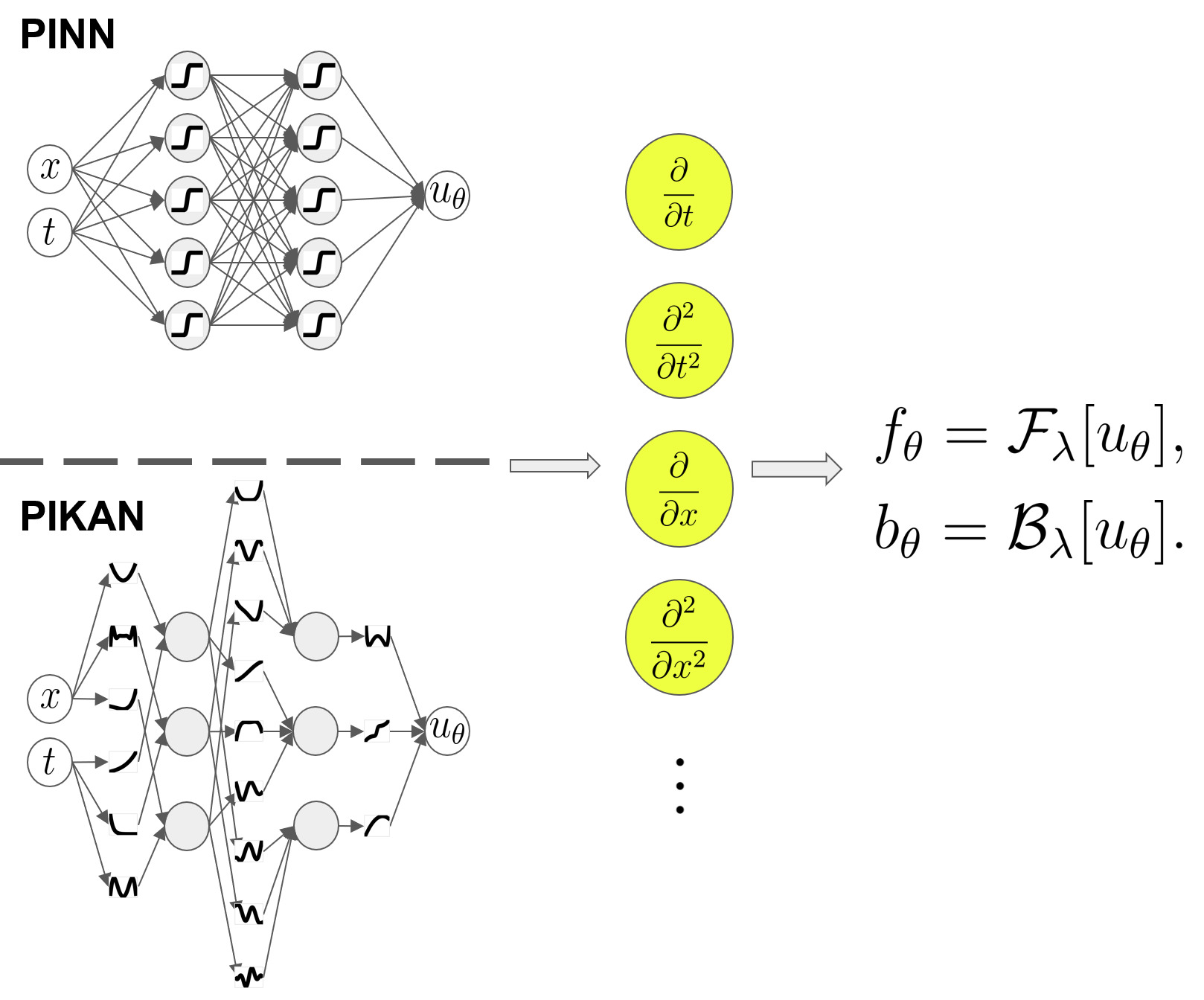}
    }
    \subfigure[Operator networks.]{
    \includegraphics[scale=.27]{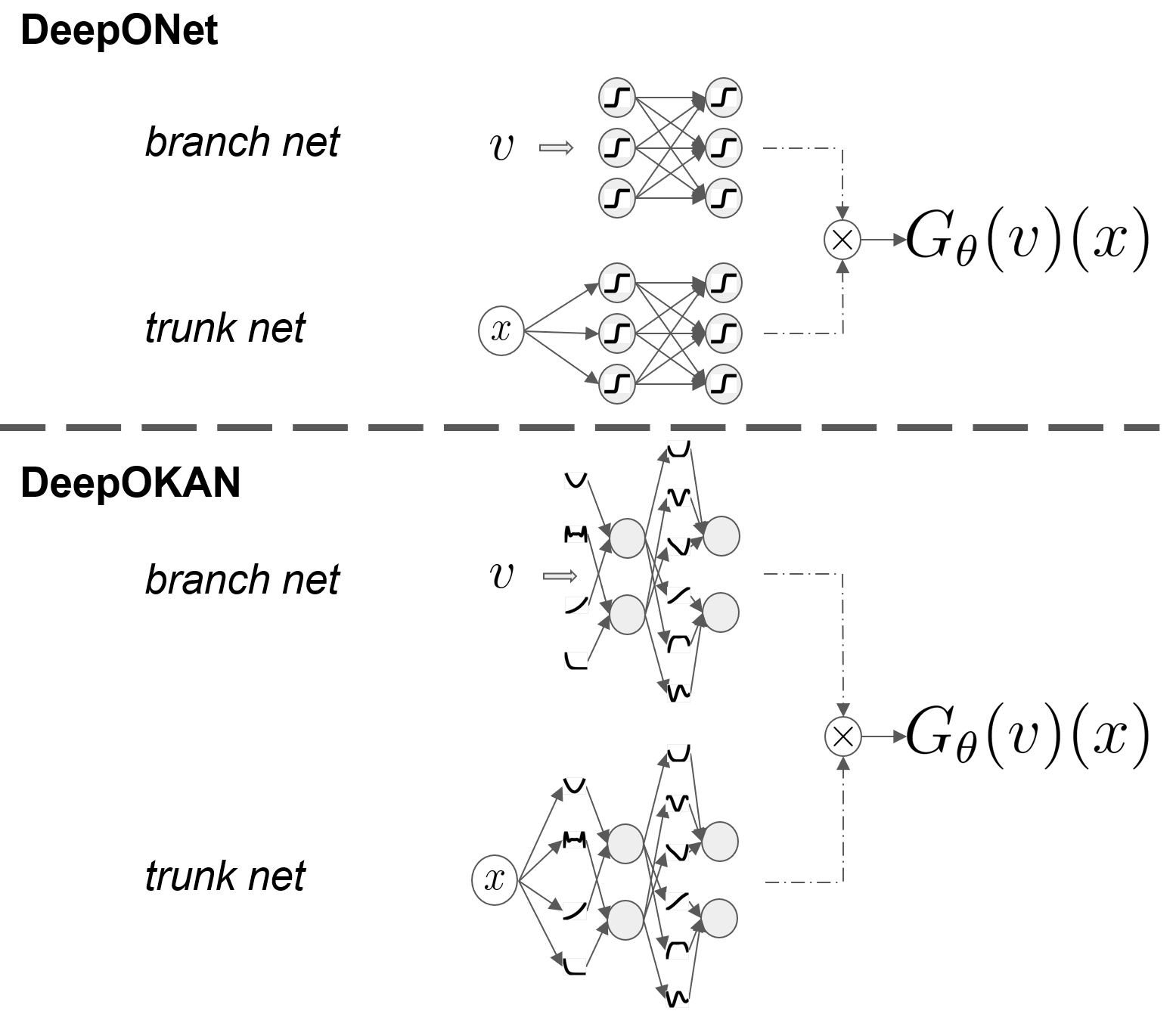}
    }
    \caption{An illustration of MLP and KAN for \textbf{(a)} differential equations and \textbf{(b)} operator networks. We choose DeepONet \cite{lu2021learning} as the representation model for operator learning. Here activation function for MLP in PINNs and DeepONets is chosen as the hyperbolic tangent only for the demonstration.}
    \label{fig:1}
\end{figure}

Consider the following nonlinear ODE/PDE:
\begin{subequations}\label{eq:problem}
    \begin{align}
        \mathcal{F}_\lambda[u](x) &= f(x), x\in\Omega, \\
        \mathcal{B}_\lambda[u](x) &= b(x), x\in\partial\Omega,
    \end{align}
\end{subequations}
where $x$ is the spatial-temporal coordinate, $u$ is the solution, $\lambda$ is the model parameter, $f$ is the source term, $b$ is the boundary term, and $\mathcal{F}$ and $\mathcal{B}$ are general nonlinear differential and boundary operators, respectively. In the literature \cite{karniadakis2021physics, psaros2023uncertainty}, there are, generally, two ways to solve \eqref{eq:problem} with modern machine learning techniques. One is approximating the solution $u$ with parameterized model, $u_\theta$ where $\theta$ denotes the parameter, constructing physics-informed loss function via automatic differentiation, and finding $\theta$ such that the loss function is minimized \cite{raissi2019physics, cai2021physics, mao2020physics, YANG2021109913, meng2020ppinn,  zou2023hydra, zou2024correcting, zhang2024discovering, chen2024leveraging, chen2023leveraging}. Representative model is physics-informed neural networks (PINNs) \cite{raissi2019physics}. In this work, we refer to it as the \textit{neural differential equation}. Another one is learning the solution operator, which maps $f, b$ and/or $\lambda$ to $u$, using NNs. Representive models are deep operator networks (DeepONets) \cite{lu2021learning} and Fourier neural operators (FNOs) \cite{li2020fourier}. We refer to them as \textit{neural operators}. The main difference between neural differential equations and neural operators is that the former targets at solving one specific ODE/PDE, in which the training of NNs gives an approximated solution that maps point to point, while the latter aims to solve a family of ODEs/PDEs, in which NNs map functions to functions.

\subsection{Physics-informed neural networks (PINNs)}

The PINN method \cite{raissi2019physics} solves the problem involving \eqref{eq:problem} by modeling the sought solution with a NN, denoted by $u_\theta$, and then modeling $f$ and $b$ with $\mathcal{F}[u_\theta]$ and $\mathcal{B}[u_\theta]$ via automatic differentiation, respectively. The differential equation is explicitly encoded by constructing the physics-informed loss function as follows:
\begin{equation}\label{PINN_loss}
    \begin{aligned}
        \mathcal{L}(\theta) = \frac{w_u}{N_u} \sum_{i=1}^{N_u} ||\alpha_i(u_\theta(x_i^u) - u_i)||^2 + \frac{w_f}{N_f}\sum_{j=1}^{N_f}||\alpha_j(\mathcal{F}_\lambda[u_\theta](x_j^f) - f_j)||^2 + \\
        \frac{w_b}{N_b}\sum_{l=1}^{N_b}||\alpha_l(\mathcal{B}_\lambda[u_\theta](x_l^b) - b_l)||^2,
    \end{aligned} 
\end{equation}
where $w_u, w_f, w_b$ are belief weights for different terms in the loss function, $||\cdot||$ is the $l^2$ norm for finite-dimensional vectors, $\{x_i^u, u_i\}_{i=1}^{N_u}, \{x_j^f, f_j\}_{j=1}^{N_f}, \{x_l^b, b_l\}_{l=1}^{N_b}$ are data for $u, f, b$, and $\alpha_i,\alpha_j$ and $\alpha_l$ are local weights (such as residual-based attention weights) that balance the loss contribution of the training points $i$, $j$ and $l$, respectively. We note that $N_u=0$ when the ODE/PDE is solved with known $\lambda$, which is often referred to as the \textit{forward problem} \cite{raissi2019physics, meng2020ppinn} in the literature, while $N_u\neq 0$ when $\lambda$ is unknown, which is referred to as the \textit{inverse problem} \cite{shukla2020physics, shukla2021physics}.

\paragraph{Residual-Based Attention}

  One of the inherent challenges in training neural networks is that the residuals (i.e., point-wise errors) can get overlooked when calculating the cumulative loss function (i.e., summation or mean of the residuals). To address this issue, several studies proposed scaling the loss terms using local multipliers \cite{mcclenny2023self,anagnostopoulos2024residual}. Local multipliers such as residual-based attention (RBA) weights \cite{anagnostopoulos2024residual} or self-adaptive weights \cite{mcclenny2023self} have shown a remarkable performance in physics-informed neural networks and other supervised learning tasks. These weights balance the contribution of specific training points within each loss term inducing a residual homogeneity \cite{anagnostopoulos2024learning}. RBA weights are based on the exponentially weighted moving average of the residuals. Thus, since the loss residuals contain information about the high error regions, the obtained multipliers work as an attention mask that helps the optimizer focus on capturing the spatial or temporal characteristics of the specific problem \cite{anagnostopoulos2024residual}.

  The update rule for RBA for any training point $i$ on iteration $k$ is given by:

  \begin{equation}
  \alpha_i^{k+1} \leftarrow (1-\eta^*)\alpha_i^{k}+\eta^* {\frac{\left| e_i \right|}{\;\,\,\lVert e \rVert_{\infty}}}, \ \ i \in \{0, 1, ..., N\},
  \label{Update_RBA}
  \end{equation}
      
    \noindent where $N$ is the number of training points, $e_i$ is the residual of the respective loss term for point $i$ and $\eta^*$ is a learning rate.  This is a convergent linear homogeneous recurrence relation that bounds our RBA between zero to one ($\alpha\in [0,1]$).

\subsection{Neural operators (NOs)}

NOs solve \eqref{eq:problem} by approximating the solution operator, denoted as $G_\theta$, using NNs from data \cite{lu2021learning, li2020fourier, lu2022comprehensive, psaros2023uncertainty, zou2024large, shukla2024deep}. Unlike neural differential equations in which the NN maps point to point, i.e. the spatial-temporal coordinate to the value of the function evaluated at this coordinate, NOs address mappings from functions to functions, e.g. the mapping from the source term $f$ to the sought solution $u$, and can be used as fast solvers for forward problems and physics encoders \cite{li2020fourier, meng2022learning, zou2024neuraluq, zou2023uncertainty} for inverse problems. We denote the input function as $v$ and the output function as $u$, and then the loss function can be formulated as follows:
\begin{equation}
    \mathcal{L}(\theta) = \frac{1}{N} \sum_{i=1}^N\frac{1}{N_u^i}\sum_{j=1}^{N^i_u} ||G_\theta(\mathbf{v}_i)(x_i^j) - u_i^j||^2, 
\end{equation}
where $\{\mathbf{v}_i, \{x_i^j, u_i^j\}_{j=1}^{N_u^i}\}_{i=1}^N$ are the data for training the NO. Here, $N$ denotes the number of paired data for the input function $v$ and the output function $u$, $\mathbf{v}$ denotes the finite representation of $v$, and $N_u^i, i=1,...,N$ is the number of measurements for the $i$th data for $u$, which is denoted by $x_i^j, u_i^j, j=1,..., N_u^i$.

\subsection{Representation Models}

\subsubsection{Multilayer Perceptron (MLP)}

The output $y$ of a Multilayer Perceptron  (MLP) can be described by the following nested formulation, where \( \sigma \) denotes the activation function, \( W^{(l)} \) and \( b^{(l)} \) are the weights and biases of the \( l \)-th layer, respectively:

\[
y(\bm{x})=\sigma\left(W^{(L)} \sigma\left(W^{(L-1)} \ldots \sigma\left(W^{(1)} \textbf{x} + b^{(1)}\right) \ldots + b^{(L-1)}\right) + b^{(L)}\right)
\]

In this formula, \( \textbf{x} = (x_1, x_2, \cdots) \) represents the input vector, and \( L \) is the number of layers. Each layer's output serves as the input for the next layer, culminating in the final output \( y \). This structure, combined with sufficiently many neurons and the right choice of activation function, allows MLPs to approximate virtually any continuous function on compact subsets of \( \mathbb{R}^n \), as stated by the Universal Approximation Theorem \cite{hornik1989multilayer}. This theorem underpins the ability of neural networks to model complex, nonlinear relationships. 

The combination of physics-informed machine learning and MLPs is called physics-informed neural networks (PINNs). Based on this definition, we can define the specific number of parameters ($|\theta|$) of PINNs as follows:

\begin{equation}
    |\theta|_{PINN}=H[I+(n_l-1)H+O]\sim \mathcal{O}(n_lH^2)\\
    \label{Params_PINN}
\end{equation}
\noindent where $I$ and $O$ are the numbers of inputs and outputs, $n_l$ is the number of hidden layer and $H$ is the number of neurons per hidden layer.

\subsubsection{Kolmogorov-Arnold networks (KANs)}

Kolmogorov-Arnold networks (KANs) are a novel type of neural network inspired by the Kolmogorov-Arnold representation theorem. This theorem states that any multivariate continuous function $f(\bm{x}) = f(x_1, x_2, \dots)$ on a bounded domain can be represented as a finite composition of continuous functions of a single variable, and the binary operation of addition \cite{liu2024kan}. Motivated by this theorem, \cite{liu2024kan} proposed approximating $f(\bm{x})$ as follows:

\begin{equation}
    f(\bm{x}) \approx \sum_{i_{L-1}=1}^{n_{L-1}} \phi_{L-1,i_L,i_{L-1}} \left( \sum_{i_{L-2}=1}^{n_{L-2}} \cdots \left( \sum_{i_2=1}^{n_2} \phi_{2,i_3,i_2} \left( \sum_{i_1=1}^{n_1} \phi_{1,i_2,i_1} \left( \sum_{i_0=1}^{n_0} \phi_{0,i_1,i_0}(x_{i_0}) \right) \right) \right) \cdots \right)
\label{KAN_net}
\end{equation}

The right-hand side of \eqref{KAN_net} represents a KAN ($KAN(\bm{x})$), where $L$ denotes the number of layers, $\{n_j\}_{j=0}^{L}$ are the numbers of nodes (i.e., neurons) in the $j^{th}$ layer, and $\phi_{i,j,k}$ are the univariate activation functions. The specific form of each $\phi(x)$ defines the variations among different KAN architectures.

\paragraph{Vanilla KAN (PIKAN)} In the original implementation \cite{liu2024kan} proposed defining $\phi(x)$ as a weighted combination of a basis function $b(x)$ and B-splines. In particular:
\begin{align*}
    \phi(x)=w_bb(x)+w_s\text{spline}(x)
\end{align*}
where the basis function $b(x)$ and the spline function $\text{spline}(x)$ are defined as follows:
\begin{align*}
    b(x)&=\frac{x}{1+e^{-x}}\\
    \text{spline}(x)&=\sum_ic_iB_i(x)
\end{align*}
here, $c_i$, $w_b$ and $w_s$ are trainable parameters. The splines $B_i(x)$, are characterized by the polynomial order $k,$ and the number of grid points $g$. Notice that, under this formulation, the trainable parameters define the contribution of each univariate function. In this study, we denote PIKAN as the combination of physics-informed machine learning and vanilla KANs. As described in \cite{liu2024kan}, the total number of parameters of KANs  (and PIKANs) can be quantified as follows:
\begin{equation}
    |\theta|_{PIKAN}=H[I+(n_l-1)H(k+g)+O]\sim \mathcal{O}(n_lH^2(k+g))\\
    \label{Params_PIKAN}
\end{equation}
\noindent where, $I$ and $O$ are the numbers of inputs and outputs, $n_l$ is the number of hidden layer, $H$ is the number of neurons per hidden layer, $g$ is the grid size, and $k$ is the polynomial order. 

Since the debut of the KAN in April 2024, researchers across the world have been actively exploring the development of KANs tailored for diverse applications. There are quite a few KAN variations appear for testing, see the Github page for the KANs collection \cite{awsomeKANs}. Among them, we would like to mention following KAN variations,

\paragraph{Radial Basis Function(RBF) KANs} It is reported by \citep{li2024kolmogorov} that using the radial basis functions (RBFs) with Gaussian kernels  toapproximates the 3-order B-spline basis, in addition with layer normalization that can prevent the inputs shifting away from the domain of the RBFs, the vanila KAN can accelerate the training without loss of accuracy.

\paragraph{Wavelet KANs}
Wavelet that uses orthogonal or semi-orthogonal basis, has the capability to maintain a balance between accurately representing the underlying data structure and avoiding overfitting to the noise. It is reported by \cite{bozorgasl2024wavkan} that the wavelet KAN is able to enhance the accuracy, speedup the training, and increase the robustness compared MLPs. 
 
\paragraph{Jacobi KANs} Jacobi polynomials are orthogonal polynomials defined on the interval [-1, 1]. They are very popular at high order numerical methods for computational fluid dynamics \cite{GK_CFDbook}. The Jacobi polynomials can be calculated recursively. Note that Chebyshev polynomials and Legendre polynomials are special cases of Jacobi polynomials. Implementation of the former can be found in \cite{chebykan}, while the latter is available on \cite{OrthogPolyKANs}. It is worth noting that the the work by \cite{ss2024chebyshev} shows that the Chebyshev KAN is more efficient than the original KAN implementation and it might represent a promising step towards leveraging theoretical foundations and efficient approximation techniques in the field of machine learning. 

This study defines cPIKAN as the combination of physics-informed machine learning and Chebyshev KAN. cPIKANs do not require grid points as PIKANs which reduces the number of trainable parameters $|\theta|$ to:
\begin{equation}
    |\theta|_{cPIKAN}=H[I+(n_l-1)Hk+O]\sim \mathcal{O}(n_lH^2k)\\
    \label{Params_cPIKAN}
\end{equation}
\noindent as in the previous models, $I$ and $O$ are the numbers of inputs and outputs, $n_l$ is the number of hidden layer, $H$ is the number of neurons per hidden layer, and $k$ is the polynomial order.

%

%% file: sec_3.tex
\section{Computational experiments}\label{Sec3}
In this section, we present a series of computational experiments comparing the efficacy and accuracy of MLP and KAN-based architectures in solving SciML problems. Specifically, we focus on utilizing KAN-based architectures to solve steady and unsteady partial differential equations, perform operator regression in low and high-dimensional regimes, and explore the applicability of KAN-based architectures for solving PDEs with noisy data by utilizing the Bayesian framework. All these experiments are further benchmarked against contemporary MLP-based architectures.

\subsection{ Approximation of a discontinuous and oscillatory function }\label{func_approx_text}
Here, we compare the approximation capability of KAN and Chebyshev-KAN against the MLP architecture. To illustrate this, we select a function that includes a discontinuity and various high and low-frequency modes. The selection of this function aims to evaluate the robustness of KAN and MLP-based architectures in addressing the prevalent phenomenon of spectral bias in neural networks, as discussed in \cite{rahaman2019spectral}.
The function is expressed as follows,
\begin{equation}\label{oscillatory_function}
    y = \left\{
    \begin{aligned}
	&5+\sum_{k=1}^4 \sin (k x), ~ x < 0, \\
        &\cos (10 x), ~ x \geq 0.
    \end{aligned}
    \right.
\end{equation}

\begin{figure}
\centering
\includegraphics[trim={1cm 0cm 2cm 0cm}, clip, width=\columnwidth,keepaspectratio]{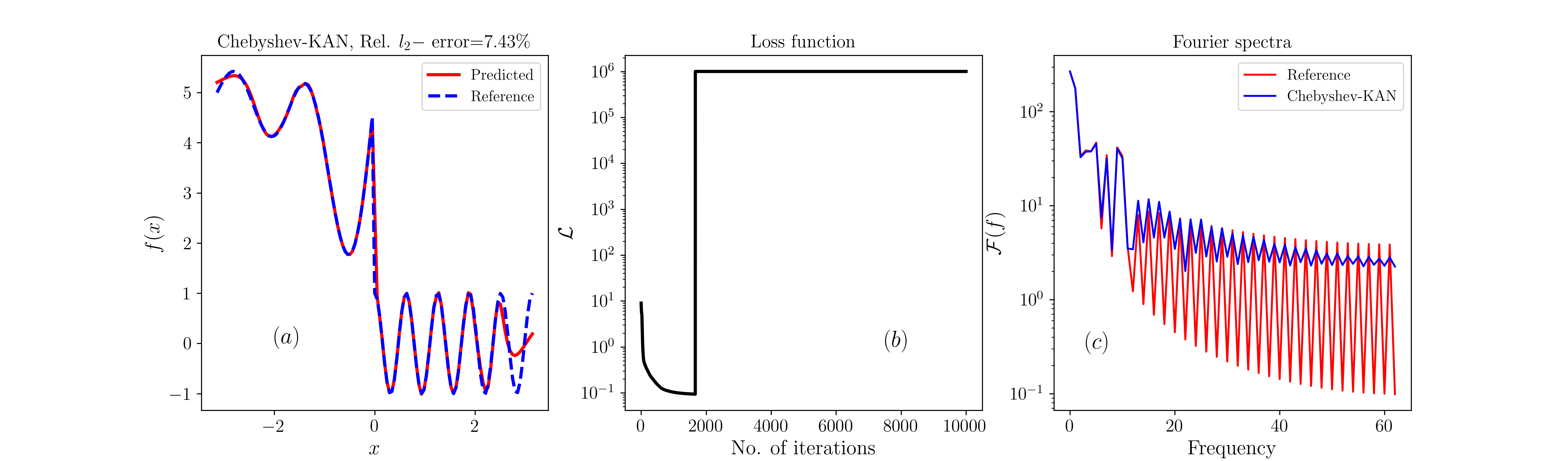}
\caption{ Expressivity of Chebyshev-KAN while approximating the function \eqref{oscillatory_function} is shown here. Subfigure (a) compares the reference and approximated functions, with the approximation by Chebyshev-KAN exhibiting a large error of 7.43\%. Subfigure (b) depicts the trajectory of the training loss, noting that training becomes unstable after the 2000$^\text{th}$ iteration, leading to NaN loss values, which are represented using a very high value of order six. Subfigure (c) compares the spectra of the reference and approximated functions, highlighting Chebyshev-KAN's failure to capture the high frequencies, resulting in the significant error.}
\label{fig:func_bad_cheb}
\end{figure}

To approximate function in \eqref{oscillatory_function}, we implement KAN \cite{li2024kolmogorov}, Chebyshev-KAN \cite{ss2024chebyshev}, Modfied-Chebyshev-KAN and MLP based architectures with hyper-parameter shown in \autoref{tab:func_approx}. To achieve this approximation, we fix a neural network architecture consisting of 2 hidden layers, each containing 40 neurons. 
However, with this architecture, the number of trainable parameters for Chebyshev-KAN and MLP are in the same order of magnitude, whereas for the KAN architecture, the number of parameters increases by an order of magnitude. 
Therefore, to ensure a fair comparison, we present the regression results using two architectures for KAN, named KAN-I and KAN-II. 
KAN-I represents the same number of layers and neurons as MLP and Chebyshev-KAN, while KAN-II represents the same number of parameters as MLP and Chebyshev-KAN. To perform these regressions for all the cases described in Section \ref{func_approx_text}, we utilize the Adam optimizer. Before discussing the main approximation results, we want to highlight the instability encountered during the training of Chebyshev-KAN architectures for approximating the function \eqref{oscillatory_function}. In \autoref{fig:func_bad_cheb}, we present the approximation of \eqref{oscillatory_function} using Chebyshev-KAN. \autoref{fig:func_bad_cheb}(a) shows the reference and approximated function plots, indicating a $l_2-$ relative error of 7.43\%. Additionally, the training becomes unstable after 2000 iterations, with the loss converging to a NaN value, as represented in \autoref{fig:func_bad_cheb}(b) by a very high double precision number. In \autoref{fig:func_bad_cheb}(c), we compare the Fourier spectrum of the reference and approximated functions, clearly showing that the network fails to accurately learn high frequencies. To address this instability, we modified the architecture by composing each Chebyshev-KAN layer with a $\tanh$ function, except for the last layer. Thus forward pass of modified Chebyshev-KAN layer with 1-hidden layer is expressed as 
\begin{align}\label{affine-ChabKan}
y = (\Phi \circ \tanh \circ \Phi)(x),
\end{align}
where $\Phi$ defines Chebyshev layer.

\begin{figure}
\centering
\includegraphics[trim={4cm 0.5cm 5cm 0cm}, clip, width=\columnwidth, keepaspectratio]{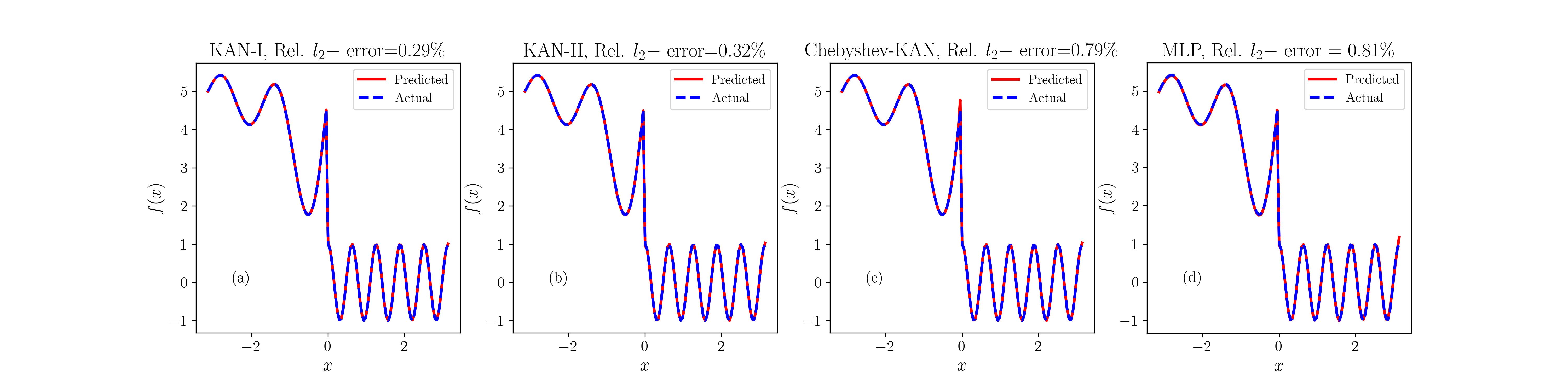}
\caption{A comparison between reference and approximated of \eqref{oscillatory_function} using  (a) KAN-I, (b) KAN-II, (c) modified Chebyshev-KAN and (d) MLP based architectures.}
\label{fig:func_approx_comp}
\end{figure}

In \autoref{fig:func_approx_comp} (a)-(d), we present the function approximation results obtained from the KAN-I, KAN-II, modified Chebyshev-KAN (equation \eqref{affine-ChabKan}), and MLP architectures, respectively. The relative $l_2-$ errors between the reference and approximated functions for the KAN-I, KAN-II, modified Chebyshev-KAN, and MLP architectures are 0.29\%, 0.33\%, 0.79\%, and 0.81\%, respectively. The expressivity of the modified Chebyshev-KAN is similar to that of the MLP architecture, although the MLP is slightly more efficient in terms of runtime. The accuracy of the KAN-I and KAN-II architectures is almost the same, despite the KAN-II having an order of magnitude fewer parameters. 

\begin{figure}
\centering
\includegraphics[trim={0cm 1cm 0cm 0cm}, clip, width=\columnwidth,keepaspectratio]{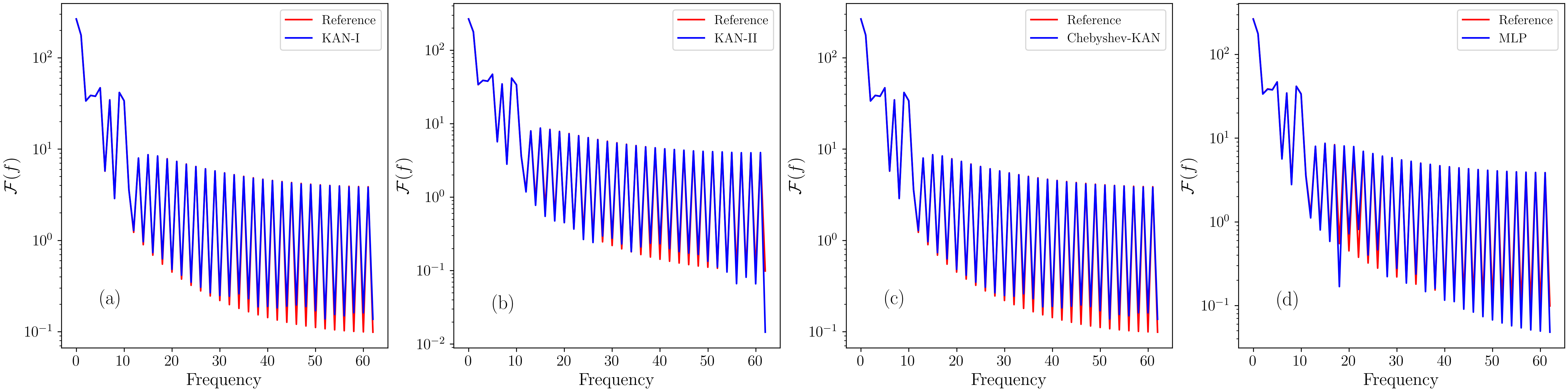}
\caption{A comparison between the spectrum of reference and approximated function obtained using (a) KAN-I, (b) KAN-II, (c) modified Chebyshev-KAN and (d) MLP architectures. Fourier spectra of approximated function obtained from all the four architectures are in very good agreement with the reference one. The long tail of oscillation represents the discontinuity present in the function \eqref{oscillatory_function}.}
\label{fig:spectra_plot}
\end{figure}

\begin{figure}
 \centering
 \includegraphics[width=0.8\textwidth]{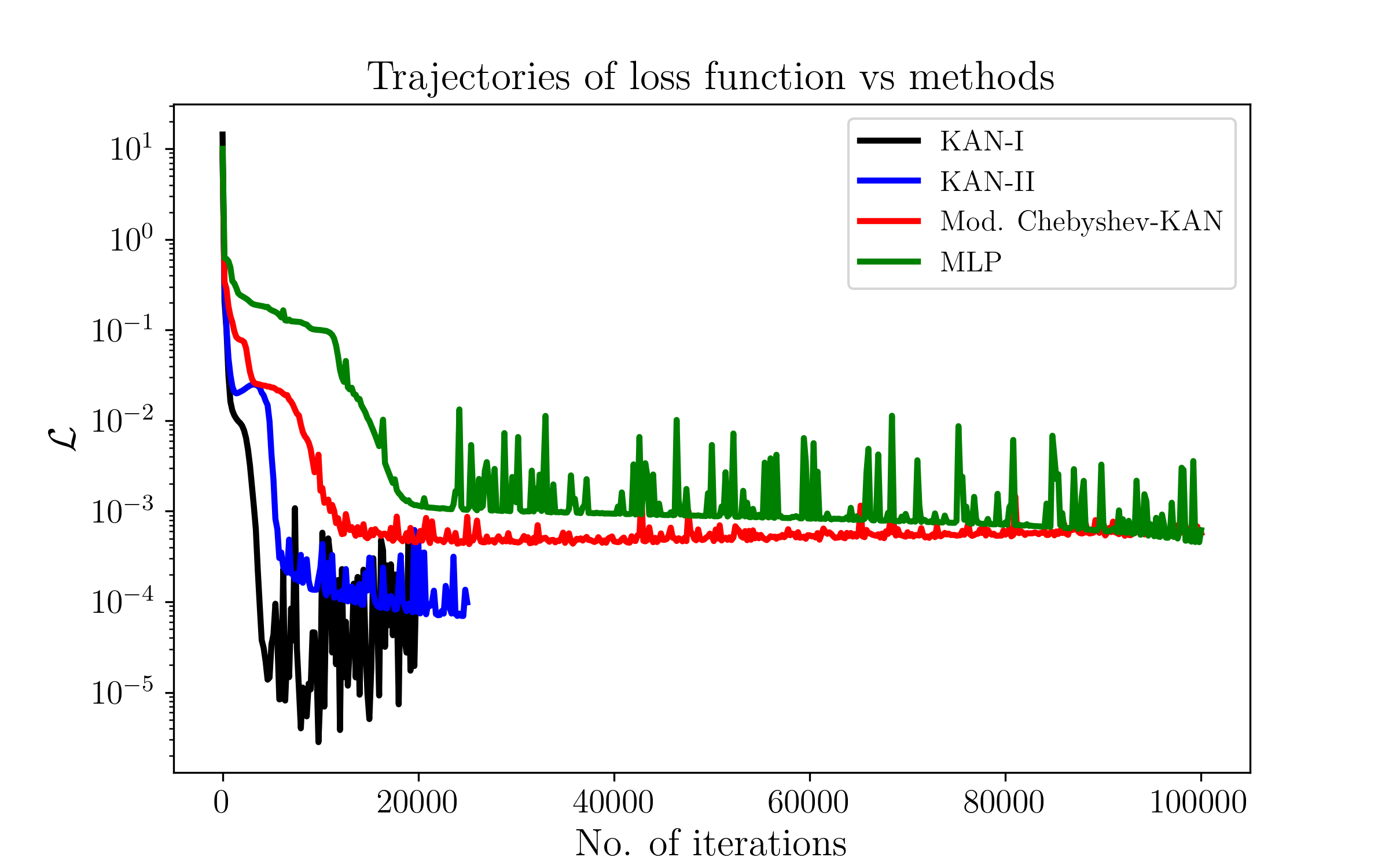}
 \caption{Loss functions for function approximation}
 \label{fig:loss_func_approx}
\end{figure}

In \autoref{fig:spectra_plot}(a)-(d), we display the Fourier spectrum of the reference and approximated functions obtained from the KAN-I, KAN-II, modified Chebyshev-KAN, and MLP-based architectures. All four architectures successfully captured all frequencies and exhibited excellent agreement with the reference spectra. In \autoref{fig:loss_func_approx}, we display the trajectories of convergence loss for the KAN-I, KAN-II, modified Chebyshev-KAN, and MLP architectures. We conducted the training for 100,000 iterations for Chebyshev-KAN and MLP due to their efficiency, whereas for KAN-I and KAN-II, the training was halted at 20,000 and 25,000 iterations, respectively, as the loss values for all four architectures converged to almost identical values. Notably, the rate of convergence for KAN-I and KAN-II is steeper compared to the modified Chebyshev-KAN and MLP architectures.

\begin{table}[h]
\centering
\scalebox{0.75}{
\begin{tabular}{|c|c|c|c|c|}
\hline
 Methods & No. of parameters &  Degree of polynomial  & Rel. $l_2-$ error & Time: $\text{ms}/\text{iter}$. \\
\hline
KAN-I \autoref{fig:func_approx_comp}a & 37041 & 3 & 0.29\% & 1586.15  \\
\hline
KAN-II \autoref{fig:func_approx_comp}b & 4317 & 3 & 0.32\% & 182.22   \\
\hline
Chebyshev-KAN \autoref{fig:func_bad_cheb} & 4352 & 3 & 7.43\% & 2.34  \\
\hline 
Modified Chebyshev-KAN \autoref{fig:func_approx_comp}c & 4352 & 3 & 0.79\% & 2.30  \\
\hline 
MLP \autoref{fig:func_approx_comp}d & 3401 & - & 0.81\% & 1.30 \\
\hline
\end{tabular}
}
\caption{Hyperparameters of networks for approximating the function \eqref{oscillatory_function} for results show in \autoref{fig:func_bad_cheb}. Here KAN-I and KAN-II repreesnts the original KAN \cite{li2024kolmogorov} however they only differ in number of parameters. The time per iteration is GPU time, measured on an Nvidia's GeForce RTX-3090 equipped with 24 GB of memory.  }
\label{tab:func_approx}
\end{table}


\subsection{Structure preserving  Dynamical System: Hamiltonian neural network (HNN) vs Hamiltonian Chebyshev-KAN (HcKAN)}
This study investigates whether Chebyshev-KAN neural networks can effectively predict the phase space of a dynamical system while preserving its energy (Hamiltonian). Traditional Multilayer Perceptrons (MLPs) struggle with this task because they lack the necessary inductive biases to guide their learning. \cite{greydanus2019hamiltonian} proposed incorporating the Hamiltonian into the training process of MLPs to address this limitation. To showcase the potential of Chebyshev-KAN for such systems, we will use a simple example: an ideal mass-spring system. The Hamiltonian for this system is given by:\cite{garg2019hamiltonian} 
\begin{align}\label{HNN-ms}
H(p,q) = \frac{1}{2}k q^2 - \frac{p^2}{2m}
\end{align}

\begin{figure}
\centering
\includegraphics[trim={0cm 0cm 0cm 0cm}, clip, width=\columnwidth,keepaspectratio]{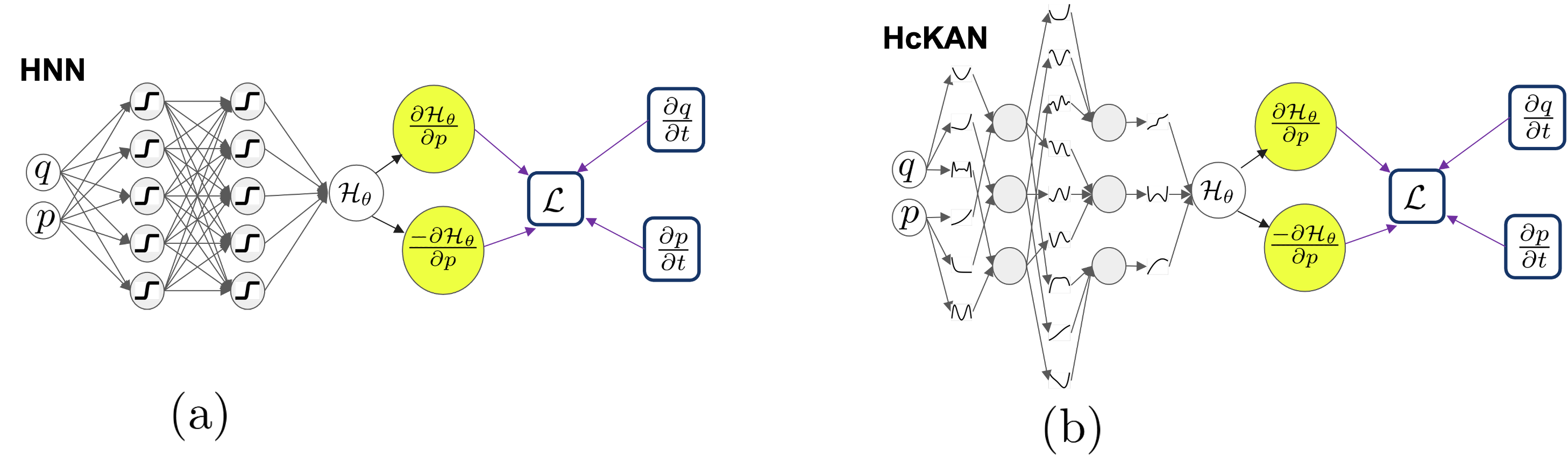}
\caption{Architectures of (a) HNN \cite{greydanus2019hamiltonian} and (b) HcKAN used for forcasting the state of the dynamical system defined by \eqref{HNN-ms} }
\label{fig:hcKAN architecture}
\end{figure}

The architecture of HNN and HcKAN is shown in \autoref{fig:hcKAN architecture}(a) and \autoref{fig:hcKAN architecture}(b), respectively. We modified the HNN in \autoref{fig:hcKAN architecture}(a) by replacing the MLP layer with with Chebyshev-KAN layer and perform the training for both HNN and HcKAN architecture by minimizing the following loss function

\begin{align}\label{loss_HKAN}
\mathcal{L}=\left\|\frac{\partial \mathcal{H}_\theta}{\partial \mathbf{p}}-\frac{\partial \mathbf{q}}{\partial t}\right\|_2+\left\|\frac{\partial \mathcal{H}_\theta}{\partial \mathbf{q}}+\frac{\partial \mathbf{p}}{\partial t}\right\|_2,
\end{align}
where $p$ and $q$ are position and momentum of the system described by equation \eqref{HNN-ms}. 

\begin{table}[h]
\centering
\begin{tabular}{|c|c|c|c|c|}
\hline
 Methods & No. of parameters &  Degree of polynomial  & N\_{train} & Time: $\text{ms}/\text{iter}$. \\
\hline
HNN \autoref{fig:hcKAN architecture}a & 4417 & - & 40 & 1.53 \\
\hline
HcKAN \autoref{fig:hcKAN architecture}b & 920 & 3 & 60 & 2.97 \\ 
\hline
\end{tabular}
\caption{Hyperparameters used for training HNN and HcKAN The time per iteration is GPU time, measured on an Nvidia's GeForce RTX-3090 equipped with 24 GB of memory. }
\label{tab:hnn}
\end{table}

\begin{figure}
\centering
\includegraphics[trim={4cm 0cm 5cm 0cm}, clip, width=\columnwidth,keepaspectratio]{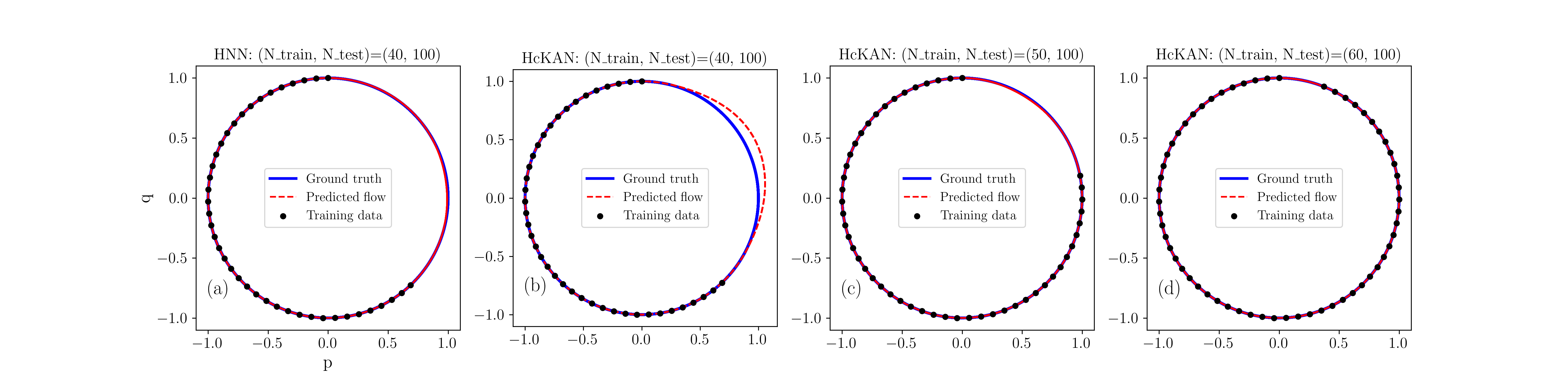}
\caption{ Learned vs actual state space of ideal mass-spring system shown by equation \eqref{HNN-ms} . (a) shows the comparison between actual and learned state space $(p, q)$ by HNN \cite{greydanus2019hamiltonian} by using [N\_train, N\_test] = [40, 100], N\_train and N\_test are number of training and testing samples. (c), (d), and (e)   represent the learned and predicted state space using [N\_train, N\_test] = [40, 100], [N\_train, N\_test] = [40, 100], and [N\_train, N\_test] = [40, 100], respectively. The bliack solid circle represent the training sample however blue and dashed red lines show the actual and predicted $p$ and $q$, respectively.}
\label{fig:hamiltonian_dynamics}
\end{figure}

\begin{figure}
\centering
\includegraphics[trim={0cm 0cm 0cm 0cm}, clip, width=\columnwidth,keepaspectratio]{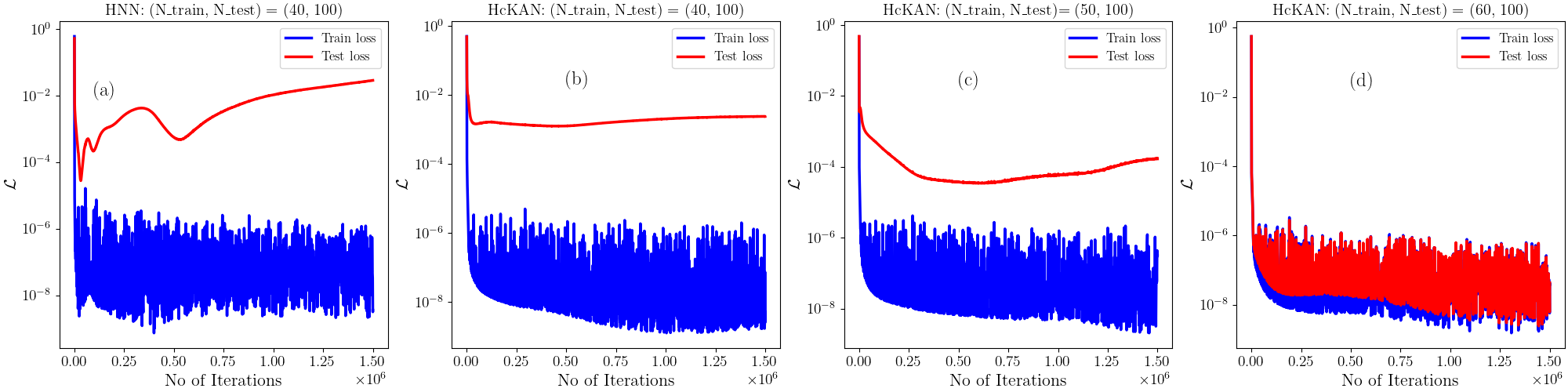}
\caption{ Train and test Loss function showing the convergence history of (a) HNN, (b) HcKAN: [N\_train, N\_test]=[40, 100], (c) HcKAN: [N\_train, N\_test]=[50, 100] and (d) (b) HcKAN: [N\_train, N\_test]=[60, 100]. }
\label{fig:loss_HNN}
\end{figure}

The hyperparameters used to train HNN and HcKAN are  provided in \autoref{tab:hnn}. To train these networks, we use the Adam Optimizer with static learning rate of 1e-3. \autoref{fig:hamiltonian_dynamics}(a)-(d) showcases how HNN and HcKAN models can learn the state space (p,q) of the ideal mass-spring system defined by equation \ref{HNN-ms}. Additionally, Figure~\ref{fig:loss_HNN}(a)-(d) present the corresponding convergence history of these models during the training process. Panel (a) of \autoref{fig:hamiltonian_dynamics} focuses on the state space learned by HNN using 40 training samples and shows good agreement between the predicted and actual state space, indicating the model's strong extrapolation capability. This observation is further supported by the convergence behavior of the training and testing loss curves in \autoref{fig:loss_HNN}(a). While the curves exhibit signs of overfitting after approximately 50,000 iterations, we selected the model with the lowest test loss for prediction purposes. In  Panel (b) of the figure \autoref{fig:hamiltonian_dynamics}  presents the state space predicted by HcKAN using the same number of training and testing samples as those employed for HNN in panel (a). Initially, we attempted to use HcKAN with the same architecture and number of parameters as the HNN (refer to \autoref{tab:hnn}). However, the training process became unstable and diverged rapidly. Reducing the number of parameters in HcKAN did not alleviate this issue. To achieve stability, we employed a modification of the Chebyshev-KAN architecture, as defined by equation~\ref{affine-ChabKan}, along with a shallower network. Panel (b) of \autoref{fig:hamiltonian_dynamics} compares the actual and predicted state space (p, q) obtained using the HcKAN model trained with 40 samples and validated with 100 additional samples. The results indicate that the extrapolation capability of HcKAN is not as strong as that of HNN and therefore have larger generalization error. The convergence history of the test loss in Figure \autoref{fig:loss_HNN}(b) further corroborates this observation. The test loss remains stagnant at a high value, indicating that the model is not generalizing well to unseen data. This improvement is further supported by the loss plots shown in Figure~\ref{fig:loss_HNN}(c) and (d). We observe a substantial decrease in test loss (by several orders of magnitude) when training with a slightly larger dataset. The execution times (runtime) for HNN and HcKAN are presented in \autoref{tab:hnn}. These measurements were conducted on an Nvidia GeForce RTX-3090 GPU. It is important to note that when HcKAN has a small number of parameters, its runtime per iteration can be higher than that of HNN. This is likely due to underutilized GPU resources; with utilization below 10\%, the latency associated with data transfer between DRAM and the processor becomes more significant than the actual computation time.

\subsection{Helmholtz Equation}

\begin{figure}[h]
 \centering
 \includegraphics[width=\textwidth]{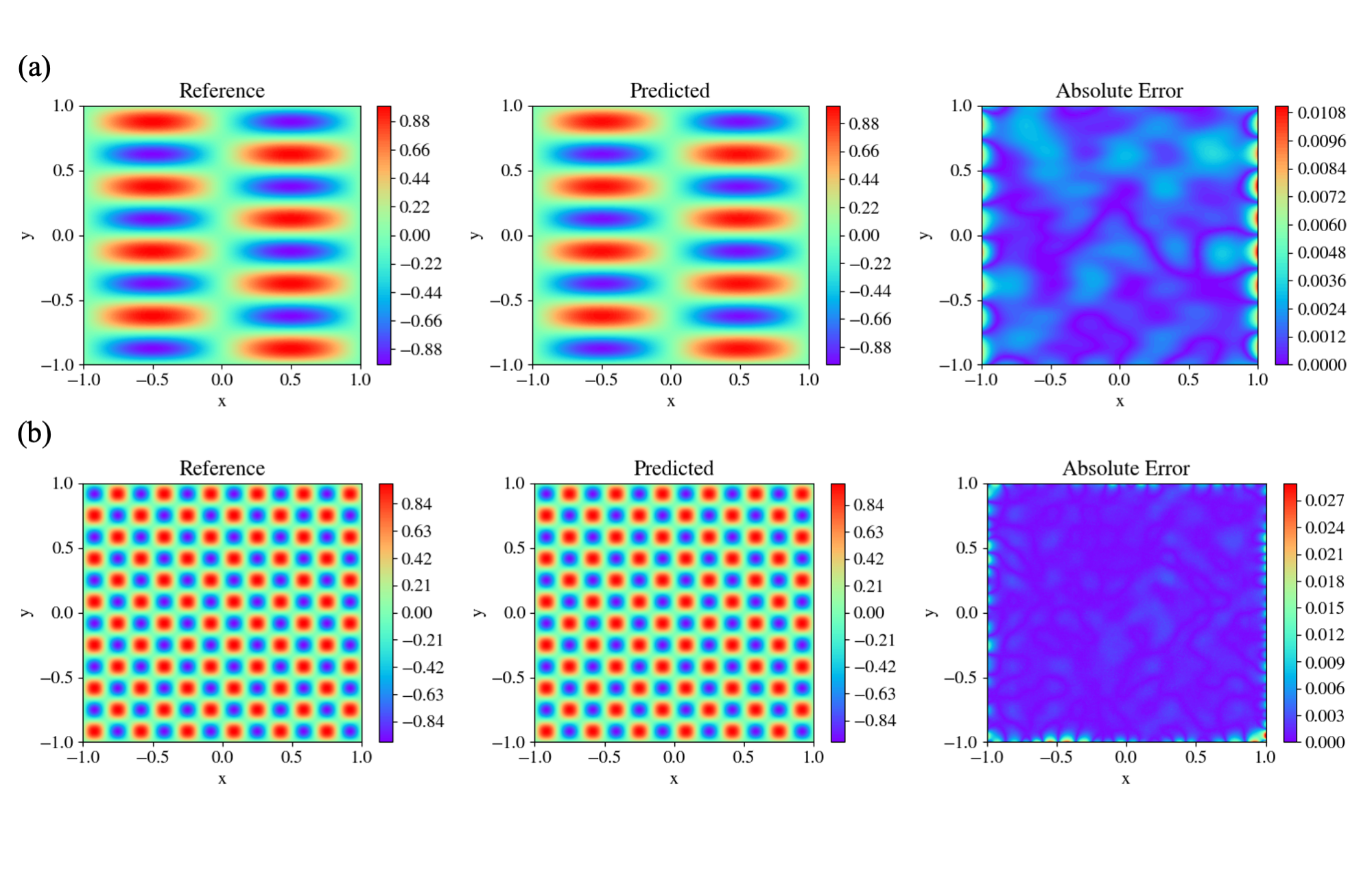}
 \caption{Reference solution with the corresponding cPIKAN+RBA prediction and the absolute error difference for different wave numbers and optimizers. (a) Helmholtz Equation with $a_1=1$ and $a_2=4$ trained with LBFGS optimizer for 1.8e3 iterations. (b) Helmholtz Equation with $a_1=a_2=6$ trained with Adam optimizer for 5.0e5 iterations.}
 \label{fig:HM_sol}
\end{figure}

The 2D Helmholtz PDE is defined as follows:
\begin{equation}
\frac{\partial^2 u}{\partial x^2} + \frac{\partial^2 u}{\partial y^2} + k^2u - q(x,y) = 0,
\label{HM_eqn}
\end{equation}

\noindent where $q(x,y)$ is a forcing term,
\begin{equation}
\begin{split}
q(x,y) = &- (a_1\pi)^2 \sin(a_1\pi x) \sin(a_2\pi y) \\
&- (a_2\pi)^2 \sin(a_1\pi x) \sin(a_2\pi y) \\
&+ k \sin(a_1\pi x) \sin(a_2\pi y),
\end{split}
\label{HM:FT}
\end{equation}

\noindent that leads to the analytical solution $u(x,y) = \sin(a_1\pi x) \sin(a_2\pi y)$ \cite{mcclenny2023self}. For this problem, the boundary conditions are expressed as:
\begin{equation}
u(-1,y) = u(1,y) = u(x,-1) = u(x,1) = 0,
\label{HM:BC}
\end{equation}

We approximate the solution of the Helmholtz equation ($a_1=1$ and $a_2=4$) with a PINN and a PIKAN by minimizing the combined loss function described in equation \eqref{HM:Loss}.
\begin{equation}
\mathcal{L} = w_{bc}\mathcal{L}_{bc} + w_{pde}\mathcal{L}_{pde},
\label{HM:Loss}
\end{equation}
\noindent here $w_{bc}$ and $w_{pde}$ are global weights that modify the contribution of the averaged loss terms for boundary conditions ($\mathcal{L}_{bc}$) and PDE residuals  ($\mathcal{L}_{pde}$) which are described as follows:
\begin{align}
\mathcal{L}_{bc} &=\langle (\alpha_{i} \cdot  \sum_{b=1}^{2}|\mathcal{R}_{i,b}|)^2\rangle_i,\\
\mathcal{L}_{pde} &= \langle (\alpha_{j} \cdot |\mathcal{R}_{j}|)^2 \rangle_j,
\end{align}
\noindent here, $\langle\cdot\rangle$ is the mean operator, $\mathcal{R}_{i,b}$ and $\mathcal{R}_{j}$ are the residuals for boundary conditions and PDE at points $i$ and $j$, respectively. $\alpha_{j}$  and $\alpha_{i}$ are RBA weights that balance the local contribution within each loss term \cite{anagnostopoulos2024residual}.

\paragraph{Parameter-based analysis}
We define suitable architectures to approximately match the number of parameters between all models. PINN has two hidden layers with 16 neurons, cPIKAN (i.e., physics-informed Chebyshev KAN)has two hidden layers with eight neurons and $k=5$, and PIKAN (i.e., physics-informed KAN) has a single hidden layer with ten neurons and  $k=g=5$. Additionally, as described in \cite{liu2024kan}, we explore the PIKAN multi-grid approach; for this case, we set $k=3$, initialize $g=5$, and divide the training process into three stages, duplicating the number of grid points every 600 iterations. We train our models by minimizing equation \eqref{HM:Loss} for $1800$ LBFGs iterations on a sample space of $51\times51$  collocation points. Following \cite{liu2024kan}; we set $w_{bc}=1$ and $w_{pde}=0.01$, which induces a biased loss function that downscales the PDE contribution. This loss function enables us to train models with few parameters in a few numbers of iterations using second-order optimizes directly. We initialize our RBA weights to one (i.e., $\alpha_i=\alpha_j=1$) and update them as described in equation \eqref{Update_RBA} with $\eta^*=1e-4$. 

\begin{figure}[H]
 \centering
 \includegraphics[width=\textwidth]{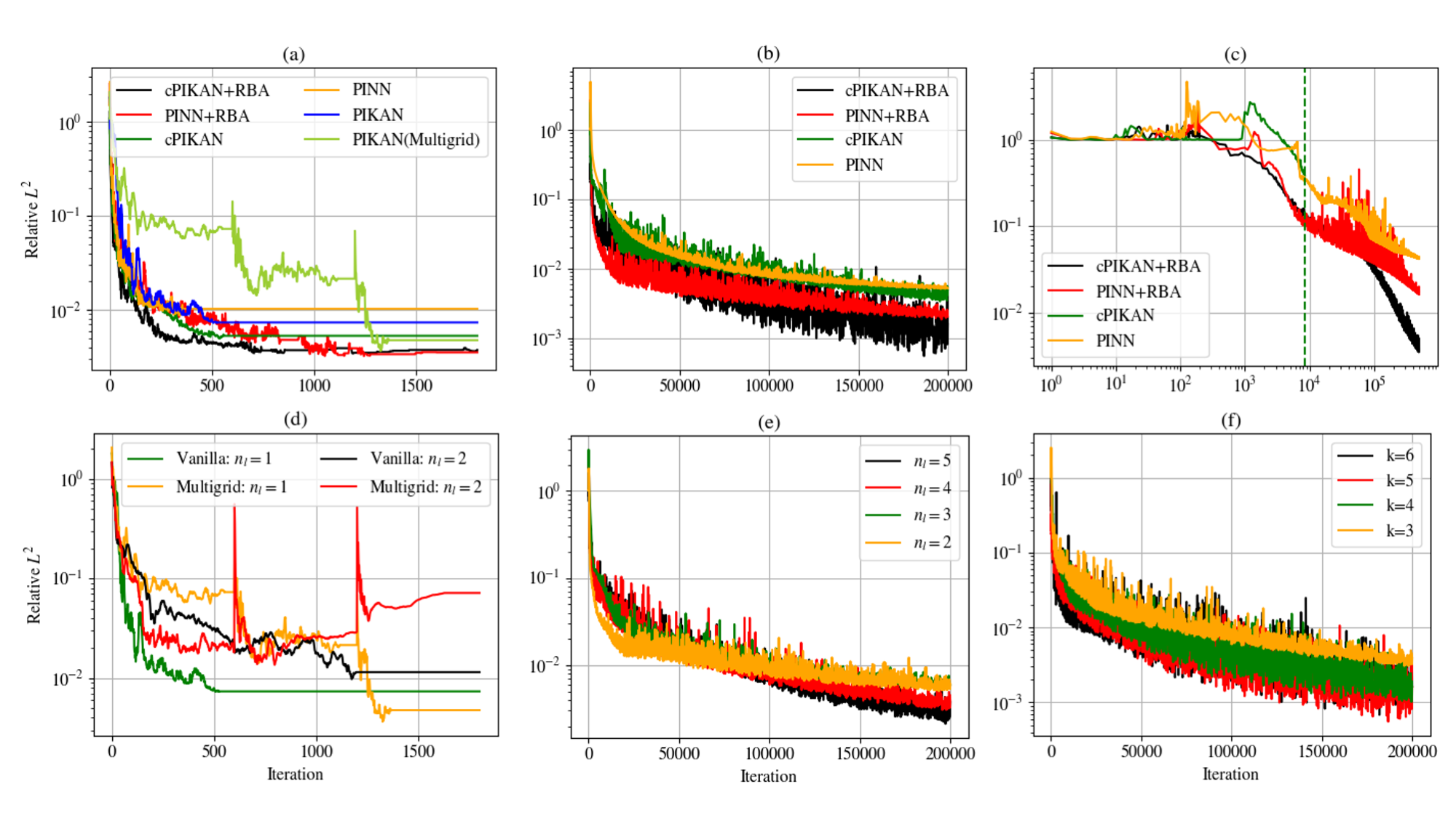}
 \caption{Relative $L^2$ convergence history. (a) Parameter-based analysis based in the original study \cite{liu2024kan} using LBFGS and global weights to downscale the PDE residuals for solving the Helmholtz equation ($a_1=1,a_2=4$). (b) Time-based analysis using ADAM optimizer with no global weights for solving the Helmholtz equation ($a_1=1,a_2=4$). (c) Complexity analysis using ADAM optimizer with no global weights for solving the Helmholtz equation ($a_1=6,a_2=6$). The green line represents the iteration where cPIKAN becomes undefined and cannot be trained further. (d) Vanilla PIKAN and multigrid PIKAN sensitivity analysis for a different number of layers for solving the Helmholtz equation ($a_1=1,a_2=4$) (e)cPIKAN+RBA ($k=3$) sensitivity analysis for a different number of layers for solving the Helmholtz equation ($a_1=1,a_2=4$). We report the number of layers up to ($n=5$) since deeper networks did not converge. (f) cPIKAN+RBA sensitivity analysis for different Chebyshev polynomial orders for solving the Helmholtz equation ($a_1=1,a_2=4$). We present the results up to degree ($k=6$) since higher orders did not converge. } 
 \label{HM}
\end{figure}

\begin{table}[h]
\centering
\begin{tabular}{|c|c|c|c|c|c|c|}
\hline
 &Method &  N. Params & Optimizer&Iterations& Relative $L^{2}$ &Time$(ms/it)$\\
\hline
 a&PINN&  304 &LBFGS&1.8e3& 1.03\%&64\\
 &PIKAN& 300 &LBFGS&1.8e3& 0.735\%&4550 \\
 &PIKAN(Multigrid)& 240-690 &LBFGS&1.8e3&0.476\%&3243\\
 &cPIKAN&  350 &LBFGS&1.8e3& 0.530\%&183 \\
 &PINN+RBA&  304 &LBFGS&1.8e3& 0.354\%&108 \\
 &cPIKAN+RBA&  350 &LBFGS&1.8e3& 0.376\%&243 \\
\hline
b&PINN&  30300 &Adam&2.0e5& 0.530\%&5.1\\
&cPIKAN&  15840 &Adam&2.0e5& 0.500\%&6.7 \\
&PINN+RBA&  30300 &Adam&2.0e5& 0.206\%&7.1 \\
&cPIKAN+RBA&  15840 &Adam&2.0e5& 0.160\%&7.4 \\
\hline
c&PINN&  82304 &Adam&5.0e5& 4.30\%&10.3\\
&cPIKAN&  20960 &Adam&5.0e5& N/A& 8.0\\
&PINN+RBA&  82304 &Adam&5.0e5& 1.72\%&11.4 \\
&cPIKAN+RBA&  20960 &Adam&5.0e5& 0.381\%&8.4 \\
\hline
\end{tabular}
\caption{Relative $L^{2}$ and computational time ($ms/it$) comparison between different models and training strategies.(a) Parameter-based analysis for solving Helmholtz Equation ($a_1=1,a_2=4$) using LBFGS optimizer and a biased loss function that downscale the PDE contribution. (b) Computation time-based comparison for solving Helmholtz Equation ($a_1=1,a_2=4$) using ADAM optimizer with an unbiased loss function. (c) Complexity-based analysis using ADAM optimizer with no global weights for solving the Helmholtz equation with a higher wave number ($a_1=a_2=6$). For the cPIKAN model, N/A represents  "not applicable" since loss became undefined after the initial iterations. Time per iteration is measured on Nvidia's GeForce RTX-3090 GPU.}
\label{Results_HM1}
\end{table}
We evaluate the model performance based on the Relative $L^2$ and training time measured in milliseconds per iteration ($ms/it$). The cPIKAN with RBA (cPIKAN+RBA) achieves a relative $L^2$ error of  0.354\%, and its prediction and corresponding point-wise error are shown in Figure~\ref {fig:HM_sol} (a). The results for the remaining methods are detailed in Table~\ref{Results_HM1}(a), and  Figure~\ref{HM}(a) shows their corresponding Relative $L^2$ convergence. Since PIKAN does not benefit from GPU parallelization, it is significantly slower than the other models; however, its performance is better than vanilla PINN. The multigrid PIKAN is faster (i.e., average of three stages) than PIKAN and outperforms cPIKAN. However, it is essential to notice that in the last stage, the number of parameters is twice as many as in the other models. For this example, cPIKAN outperforms PINN and vanilla PIKAN, and the best-performing model is PINN+RBA. However, notice that cPIKAN+RBA's final relative $L^2$ error is comparable. 
\paragraph{Computation time-based analysis} In this section, we analyze the PINN and cPIKAN model for deeper networks (i.e., four hidden layers) and a higher number of collocation points (i.e., $100\times 100$). We define the number of neurons per layer by roughly matching the PINN and cPIKAN's computational time.  In particular, we use  $100$ and $32$ neurons per hidden layer for PINN and cPIKAN, respectively. We train our models using $w_{bc}=w_{pde}=1$, which induces an unbiased loss function akin to real-world applications. To balance the contribution of each loss term, we use RBA only on the PDE residuals, initializing them to zero  (i.e., $\alpha_j=0$) and updating them interactively with $\eta^*=1e-3$  as described in equation \eqref{Update_RBA}. Following this approach, the RBAs work as global and local weights that modify the contribution of each training point iteratively by following the network residuals. We train our models for $2.0e5$ iterations using Adam optimizer \cite{kingma2014adam} with a learning rate scheduler that starts in $1e-3$ and ends in $1e-4$.

As shown in Table~\ref{Results_HM1}(b), cPIKAN marginally outperforms PINN with and without RBA. Additionally, Figure~\ref{HM}(b) shows that combining our based models with RBA accelerates their relative $L^2$ convergence. For this example, the best-performing model is cPIKAN+RBA, which achieves a relative $L^2$ error of 0.160\%. 

\paragraph{Complexity-based analysis} To increase the problem complexity, we solve the Helmholtz equation with a higher wave number (i.e., $a_1=a_2=6$). This modification induces steeper gradients in the PDE residuals, making it difficult for the neural network to approximate. For PINN, we use six hidden layers with 128 neurons per layer, while for cPIKAN, we use five layers, 32 neurons, and $k=5$. As in the previous case,  we train our model with an unbiased loss function ($w_{bc}=w_{pde}=1$) and apply RBA (initiated at zero)only in the residuals using $\eta^*=1e-3$. We update our network parameters using Adam optimizer for $5e5$ iterations with a learning rate schedule from $1e-3$ to $5e-5$.

The best-performing model reconstruction and its corresponding point-wise error are illustrated in Figure~\ref {fig:HM_sol} (b). Table~\ref{Results_HM1}(c) shows that cPIKAN+RBA significantly outperforms the other methods, achieving a relative $L^2$ error of 0.381\%. However, notice that the vanilla cPIKAN did not converge (See Figure~\ref{HM}(c)).

\subsubsection{Sensitivity Analysis}
\begin{figure}[H]
 \centering
\includegraphics[width=0.9\textwidth]{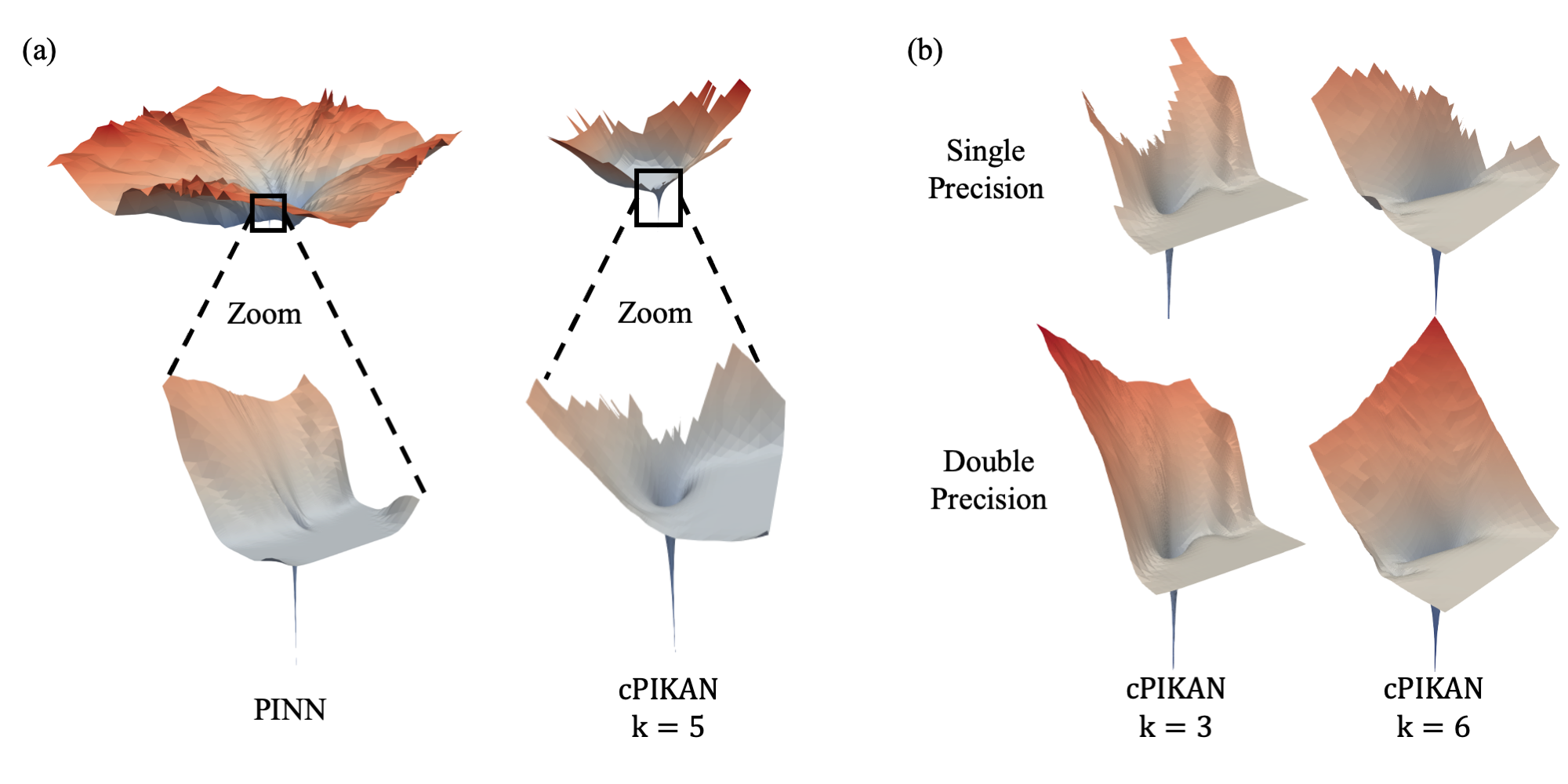}
 \caption{Los Landscapes.(a)Comparison between PINN and PIKAN ($k=5$) for solving  Helmholtz Equation ($a_1=6,a_2=6$). Notice that PIKAN's loss is not defined (empty spaces) for parameters far from the minimum, making it highly sensitive to initialization.  (b)  Comparison between PINN and PIKAN for different Chebyshev polynomial orders for solving Helmholtz Equation ($a_1=1,a_2=4$). Notice that the non-defined regions disappear when using double precision, which indicates that the problem lies within the limitations of training a model using a single precision.} 
 \label{fig:Landscapes_HM}
\end{figure}
As described in the previous sections, cPIKAN performs better than PINNs and significantly reduces the computational overhead of PIKAN. However, it induces more oscillations  (See Figure~\ref{HM}(b))and can potentially become unstable (See Figure~\ref{HM}(c)). To explore this behavior, we perform a sensitivity analysis for Helmholtz and study the influence of the number of hidden layers $n_l$ and polynomial order $k$.

First, we study the effect of $n_l$ on PIKAN ($k=5$) and PIKAN multigrid, fixing the polynomial order to $k=5$ and $k=3$ respectively. Notice that, in this case, increasing the number of layers hinders the model performance of both models. Then, we analyze the effects of $k$ (Figure~\ref{HM}(e)) and $n_l$ (Figure~\ref{HM}(e)) on cPIKANs.
 For this case, increasing $k$ or $n_l$ improves cPIKAN's performance; nevertheless, increasing these parameters makes the model unstable. In this example, using $k>6$ or $n_l>5$ causes the model's loss function to become undefined after training it for several iterations. To explore this issue, we follow\cite{li2018visualizing,wang2023solution} and plot the loss landscape for different values of $k$. An ideal landscape is smooth, continuous, and convex, which enables the optimizer to converge successfully to the global minimum. However, Figure~\ref{fig:Landscapes_HM}(a) shows that the cPIKAN landscape has empty holes near the edges, which indicate sections where the loss is not defined. Moreover, these regions take over the whole space far from the minimum, suggesting that cPIKAN models are sensitive to initialization. To further analyze this behavior, we train and plot the loss landscape using single and double pressure.
 Figure~\ref{fig:Landscapes_HM}(b) shows that, for single precision, the non-defined regions grow as we increase $k$, suggesting that the model becomes unstable for higher polynomial orders. However, it can be observed that these models can be trained, and their loss is defined when using double precision (i.e., float 64), indicating that the instability is related to the numerical approximation inherent in training a model with single precision (i.e., float 32). This indicates that it is possible to train models with higher $k$ or $n_l$ using double precision, yet this type of training increases their computational cost.

\subsection{Navier-Stokes equation}
In this section, we consider solving the following 2D steady Navier-Stokes equations with PINNs and PIKANs,
\begin{equation}\label{eq:NSE1}
    \frac{\partial u}{\partial x} +  \frac{\partial v}{\partial y} = 0,
\end{equation}

\begin{equation}\label{eq:NSE2}
    u \frac{\partial u}{\partial x} + v \frac{\partial u}{\partial y} = -\frac{\partial p}{\partial x}+\nu \big( \frac{\partial^2 u}{\partial x^2}+ \frac{\partial^2 u}{\partial y^2}\big),
\end{equation}

\begin{equation}\label{eq:NSE3}
    u \frac{\partial v}{\partial x} + v \frac{\partial v}{\partial y} = -\frac{\partial p}{\partial y}+\nu \big( \frac{\partial^2 v}{\partial x^2}+ \frac{\partial^2 v}{\partial y^2}\big),
\end{equation}
where, $(u, \,v)$ are the velocity component in $x$, $y$ direction, respectively, $p$ is the pressure, $\nu$ is the kinematic viscosity.
In particular, when the $\nu$ is small, it is recommended by \cite{He2023} to reformulate equations \eqref{eq:NSE1}-\eqref{eq:NSE3} into to the following loss functions:
\begin{equation}\label{eq:loss1}
  e_1=u\frac{\partial u}{\partial x}+v\frac{\partial u}{\partial y}+\frac{\partial p}{\partial x}-(\nu+\nu_E)(\frac{\partial^2 u}{\partial x^2}+\frac{\partial^2 u}{\partial y^2}),  
\end{equation}
\begin{equation}\label{eq:loss2}
  e_2=u\frac{\partial v}{\partial x}+v\frac{\partial v}{\partial y}+\frac{\partial p}{\partial y}-(\nu+\nu_E)(\frac{\partial^2 v}{\partial x^2}+\frac{\partial^2 v}{\partial y^2}),  
\end{equation}
~%
\begin{equation}
e_3 = \frac{\partial u}{\partial x}+\frac{\partial v}{\partial y},
\end{equation}
where $\nu_E$ is the artificial viscosity that will be determined during training. Note that $\nu_E$ is a scalar, whose construction is adapted from the entropy viscosity method (EVM) \citep{GuermondEVMJCP,wang2019jfm} for numerical stabilization in the flow simulation at a high $Re$. Specifically, $\nu_E$ can be computed from: 
\begin{equation}\label{eq:nu_e}
  \nu_E=\min(\beta_E \nu , \alpha_E \frac{|r| L^2}{U_{\infty}^2}),  
\end{equation}
where $r$ is the \emph{predicted entropy residual} and is one of the output variable of the network, 
In practice, $r$ can be inferred by following equation loss,
\begin{equation} \label{EV1}
e_4=(u-u_m)e_1+(v-v_m)e_2-r.
\end{equation}
Here $u_m$ and $v_m$ are two constants that makes $r$ be non-zero on the boundaries. In all the simulations cases of this section, $u_m=0.5$, $v_m=0.5$ are employed unless otherwise stated. 
We would like to emphasize that non-zero entropy viscosity on the wall boundary is important to improve the accuracy, which is different from the entropy viscosity in the numerical method developed in \citep{wang2019jfm}. We found that $\nu_E > 0$ in the vicinity of the boundary can result into better prediction in both PINNs and PIKANs. In equation \eqref{eq:nu_e}, $L=1$ , $U_{\infty}=1$ are the characteristic length and velocity, respectively. Moreover, $\alpha_E$ and $\beta_E$ are two tunable hyperparameters whose values may affect the inference accuracy greatly; $\alpha_E$ and $\beta_E$ can be either constant or descending throughout the training. Nonetheless, in this section, $\alpha_E=0.03$, $\beta_E=5$ are used.

\begin{table}[h]
\centering
\begin{adjustbox}{width=0.95\textwidth}
\begin{tabular}{|c|c|c|c|c|c|}
\hline
$Re$ & Method & polynomial order& Relative $L^{2}$ error $(u,v,p)$ \%  &Time $(ms/it)$&Num. of parameters \\
\hline
\multirow{12}{2em}{\textcolor{blue}{\textbf{400}}}&PINN &-& 0.25, 0.37, 2.25 & 32  & $44,283$\\
\cline{2-6}
&\multirow{3}{5em}{Chebyshev PIKAN}&3 & 1.13, 1.38, 2.34 & 61   & $4,736$\\ 
\cline{3-6}
&&5 & 0.20, 0.26, 1.63 & 81   & $7,104$\\
\cline{3-6}
&&8 & 0.18, 0.21, 1.8 & 112   & $10,656$\\
\cline{2-6}
&\multirow{2}{5em}{Legendre PIKAN}&3 &2.0, 2.3, 3.2 & 82& $4,736$  \\  
\cline{3-6} 
& &8 &0.21,0.29,1.77 & 241& $10,656$  \\
\cline{2-6}
&\multirow{2}{5em}{Jacobi PIKAN} &3 &0.31,0.43, 1.96& 120& $4,736$  \\
\cline{3-6}
& &8 &0.18,0.26, 1.67& 235& $10,656$  \\
\cline{2-6}
&\multirow{2}{5em}{Hermite PIKAN} &3 &4.1,4.9,9.7 & 84& $10,656$  \\
\cline{3-6}
& &8 &143,101,151 & 220& $10,656$  \\
\cline{2-6}
&PINN+RBA &-&0.24, 0.33, 1.83  &  34 & $44,283$\\
\cline{2-6}
&Chebyshev PIKAN+RBA &3& 0.18, 0.20, 1.78 & 92  & $4,736$ \\
\hline
\multirow{5}{2em}{\textcolor{blue}{\textbf{2000}}}&PINN &-& 18.9, 18.8, 22.8 & 63  & $44,283$\\
\cline{2-6}
&\multirow{2}{5em}{Chebyshev PIKAN}  &5& 109.2, 113.8, 132.05 & 85   & $7,104$\\
\cline{3-6}
&  &8& 105.2, 107.8, 110.23 & 127   & $10,656$\\
\cline{2-6}
&PINN+EVM &-&6.4, 5.4, 6.9  &  64 & $49,365$\\
\cline{2-6}
&Chebyshev PIKAN+EVM &8&4.2, 4.5, 8.4 & 135  & $20,736$ \\
\hline
\end{tabular}
\end{adjustbox}
\caption{Comparison on performance between PINNs and PIKANs based on different polynomials in solving 2D steady cavity flow at $Re=400$ and $Re=2000$. The PINNs, PIKANs, RBA and EVM are implemented on our NSFnet \cite{JIN2021109951}. $10^4$ training points, $9\times 10^5$ training epochs for the case of $Re=400$,  $2\times 10^4$ training points, $4\times 10^5$ training epochs for the case of $Re=2000$. The Adams optimizer \cite{kingma2014adam} is used and the training is performed on an Nvidia's RTX 4090 GPU. Note that the  coefficients of the Jacobi polynomial $P_n^{\alpha,\beta}(x)$ used in this section are: $\alpha=1$, $\beta=1$. }
\label{cavity_table}
\end{table}

We have systematically performed PINNs/PIKANs simulation of 2D steady lid-driven cavity flow at $Re=400$ and $Re=2000$. The steady cavity flow is the well-known benchmark case and is frequently used in the validation of numerical method. The details of the computational domain and boundary condition can be found in \cite{wang2023solution}. In particular, we have compared the Chebyshev, Jacobi, Legendre and Hermite polynomials based PIKANs for the cavity flow at $Re=400$ and $Re=2000$. It has been reported that vanilla PINNs can accurately infer the cavity flow at $Re=400$, but failed to obtain correct solution at $Re=2000$ \cite{wang2023solution}. In order to provide a fair comparison among PINNs and PIKANs, we have kept the same number of residual points, number of training epochs and the optimizer, i.e., the only difference between different cases lies in the network, e.g., in PINN, the multi-layer Perceptron (MLP) is used, while in Chebyshev PIKAN (cPIKAN), the Chebyshev polynomials based Kolmogorov Arnold Network (KAN) is employed. 

As shown in \autoref{cavity_table}, in total 18 cases have been tested. At $Re=400$, it could be observed that PIKANs can generate solutions with a comparable accuracy to PINNs. However, the number of trainable parameters used in PIKANs is far less than the one used in PINNs, although they can achieve the same accuracy. On the other hand, the training time for each iteration of PIKANs is four times more than the counterpart in PINNs. Among different variants of PIKANs that are differentiated by the type of polynomials, cPIKAN is the most promising architecture, in terms of the inference accuracy and GPU time of training. Moreover, with RBA, which is the technique that can dynamically and locally adjust the weights on loss function during training, both PINNs and PIKANs can further achieve better inference accuracy at the same computation time. 

From aforementioned results of $Re=400$, it could be concluded that cPIKAN is the best choice for the 2D steady cavity flow. Therefore, in the PIKAN simulation of cavity flow at $Re=2000$, only the cPIKAN is tested. However, as shown in the bottom 5 rows of Table \ref{cavity_table}, after training 40,000 epochs, the vanilla PINN manages to achieve the $l_2-$ relative error lower than 20\%, while the vanilla cPIKAN produces solutions of error higher than 100\%. However, with the help of EVM, both PINN and cPIKAN can significantly improve the inference accuracy, with the relative error lower than 7\%,  after the same number of epochs as the vanilla PINN and cPIKAN.

\begin{figure}
\centering
\includegraphics[width=0.49\textwidth,trim=120 20 120 30,clip]{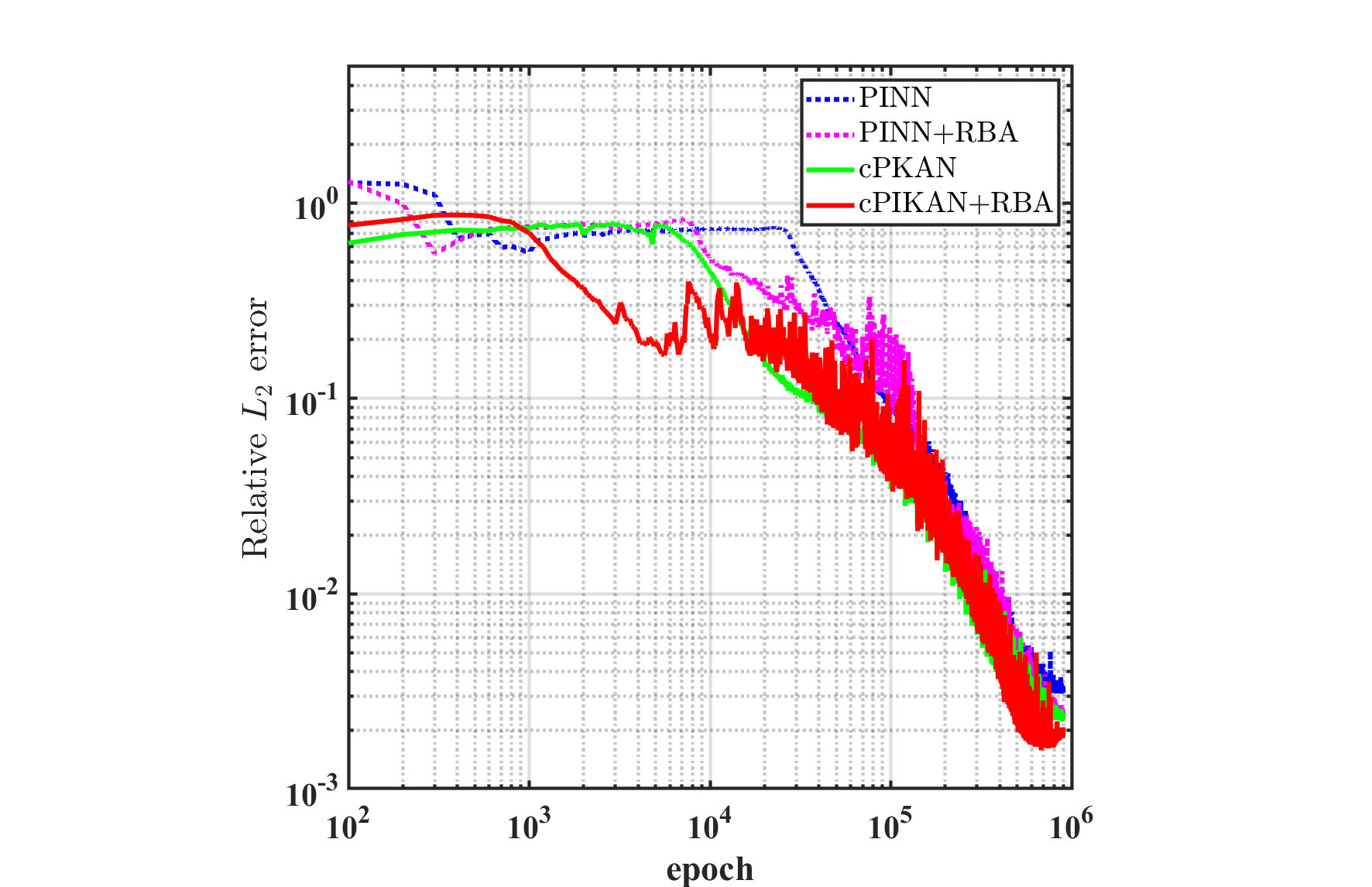}
\includegraphics[width=0.475\textwidth,trim=120 20 120 30,clip]{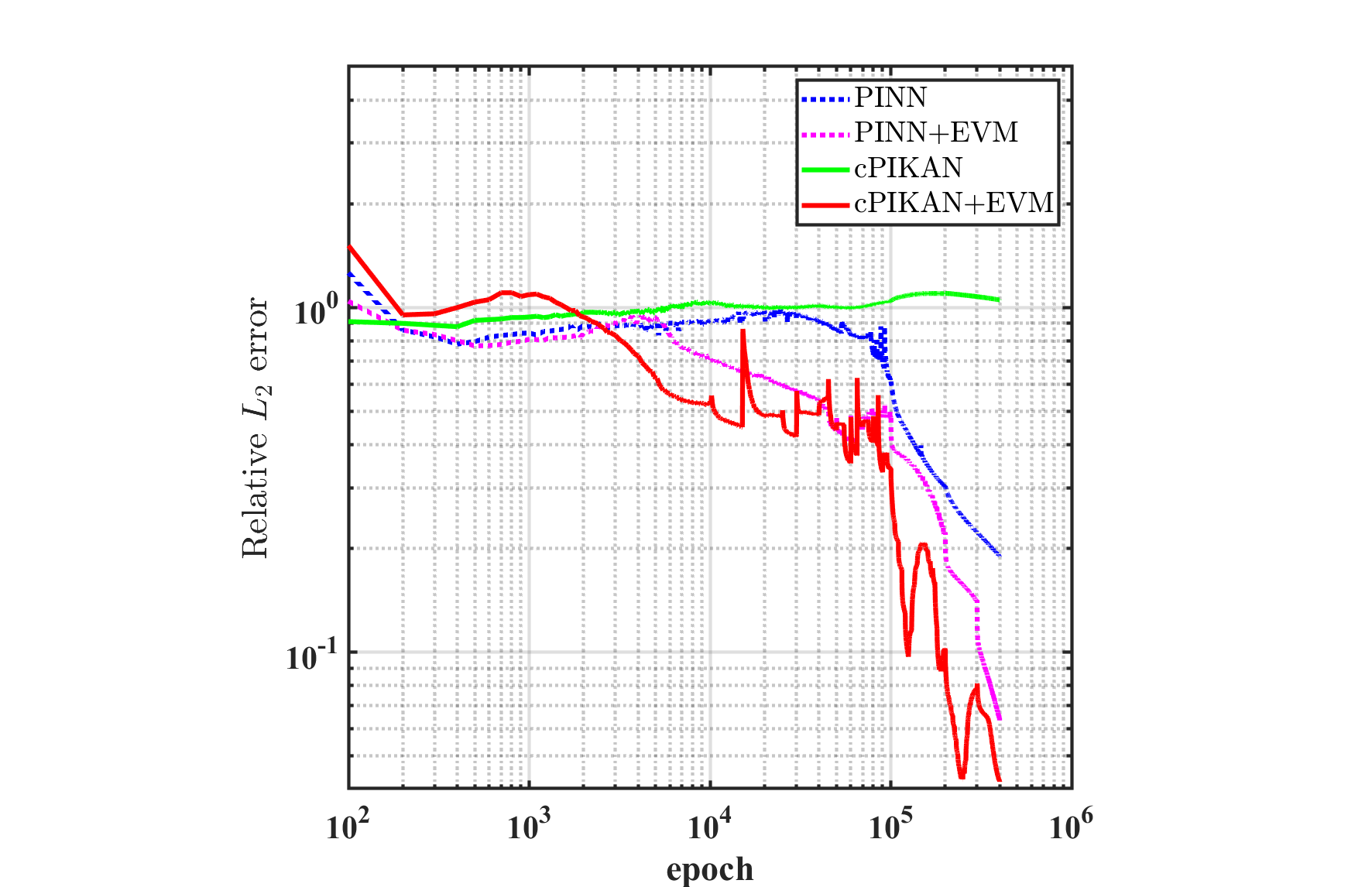}
\caption{Comparison of the error between PINN and PIKAN in solving the steady cavity flow at $Re=400$ and $Re=2000$. Note that the error is computed on a $256 \times 256$ uniform mesh, which is different from the residual points used in training. }.
\label{fig:error_cavity}
\end{figure}

The histories of relative error varying with training epochs of PINN/cPIKANs are plotted in \autoref{fig:error_cavity}. As shown in the left panel of \autoref{fig:error_cavity}, it could be observed that the RBA can speed up the both the PINN and cPIKAN training. From the right panel of \autoref{fig:error_cavity}, it can be seen that the relative error of the vanilla cPIKAN (green line) barely decays with training. However, with EVM, both PINN and cPIKAN can reduce the inference error notably, although the error history of cPIKAN exhibits more oscillations. 

\begin{figure}
\centering

\includegraphics[width=0.32\textwidth,trim=0 20 0 5,clip]{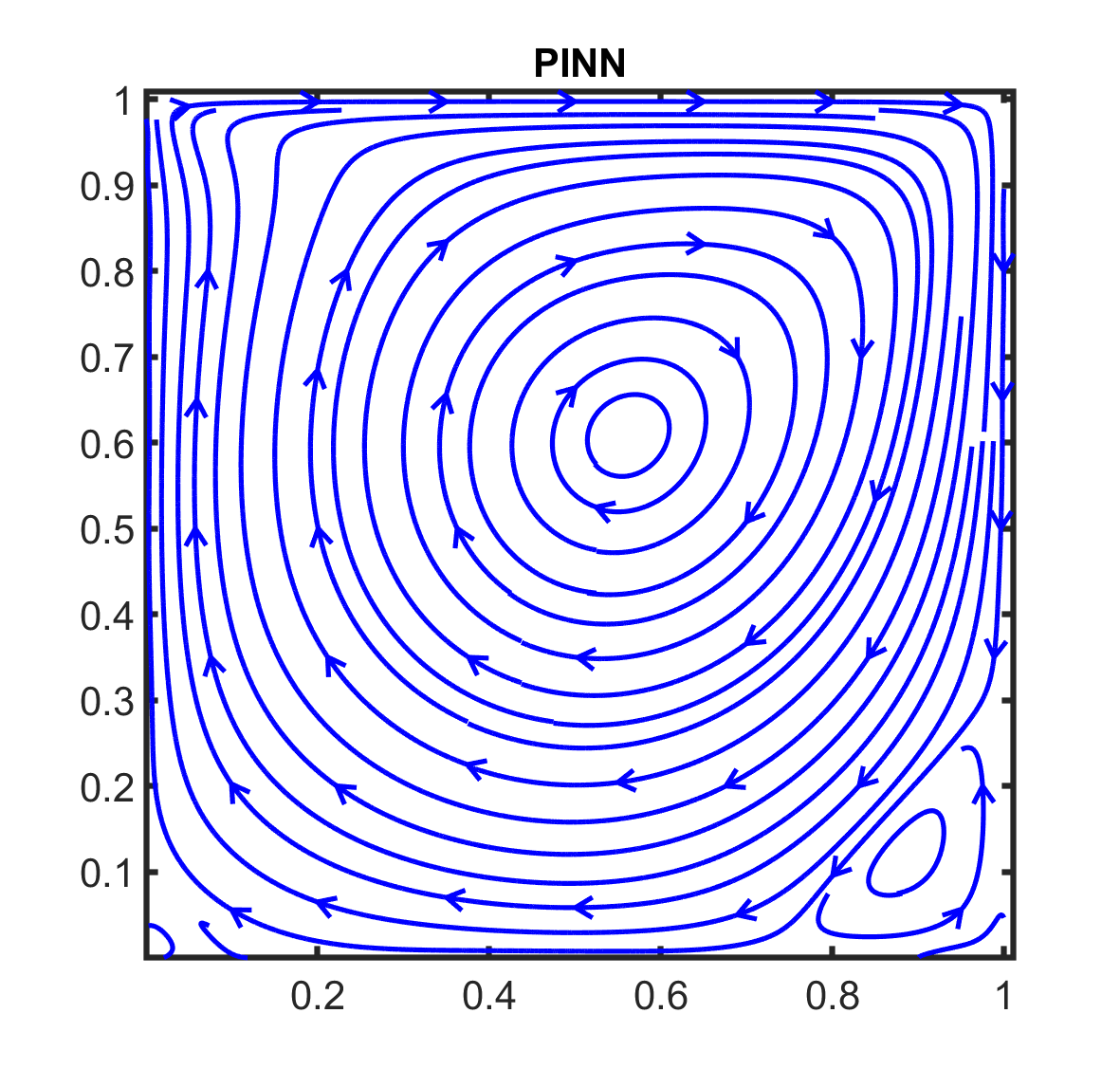}
\includegraphics[width=0.32\textwidth,trim=0 20 0 5,clip]{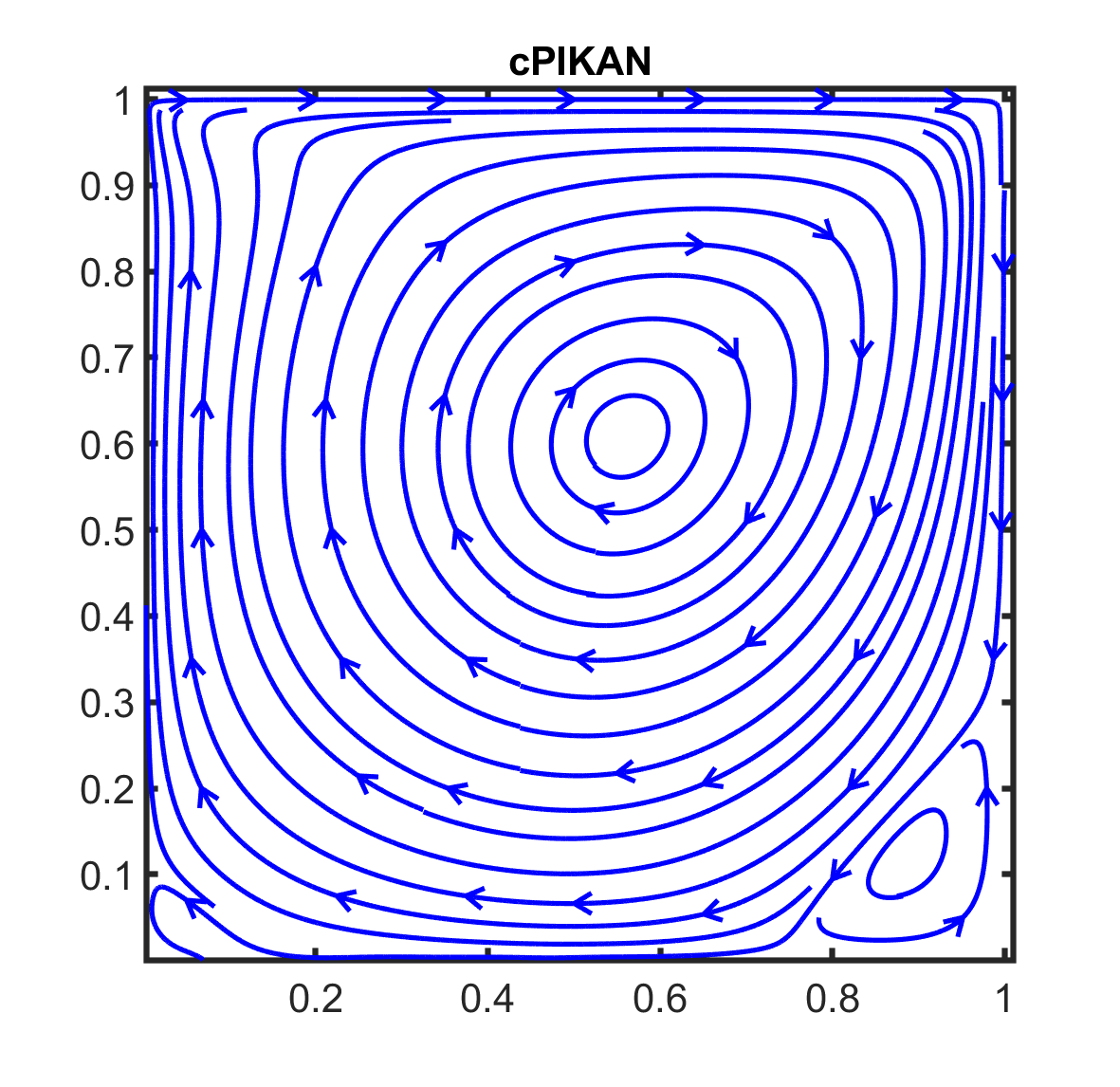}
\includegraphics[width=0.32\textwidth,trim=0 20 0 5,clip]{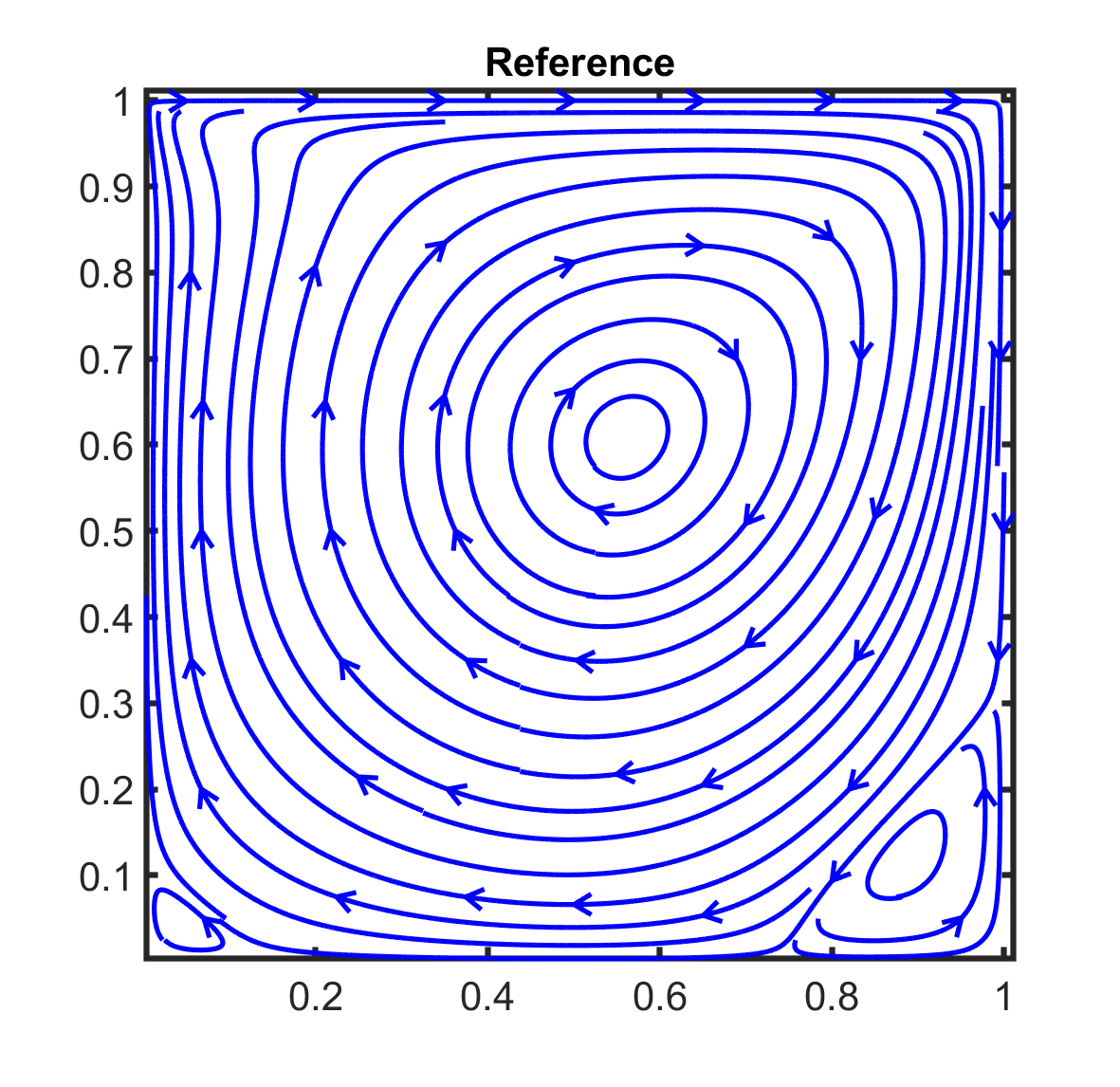}
\caption*{$Re=400$.}
~
\includegraphics[width=0.32\textwidth,trim=0 20 0 5,clip]{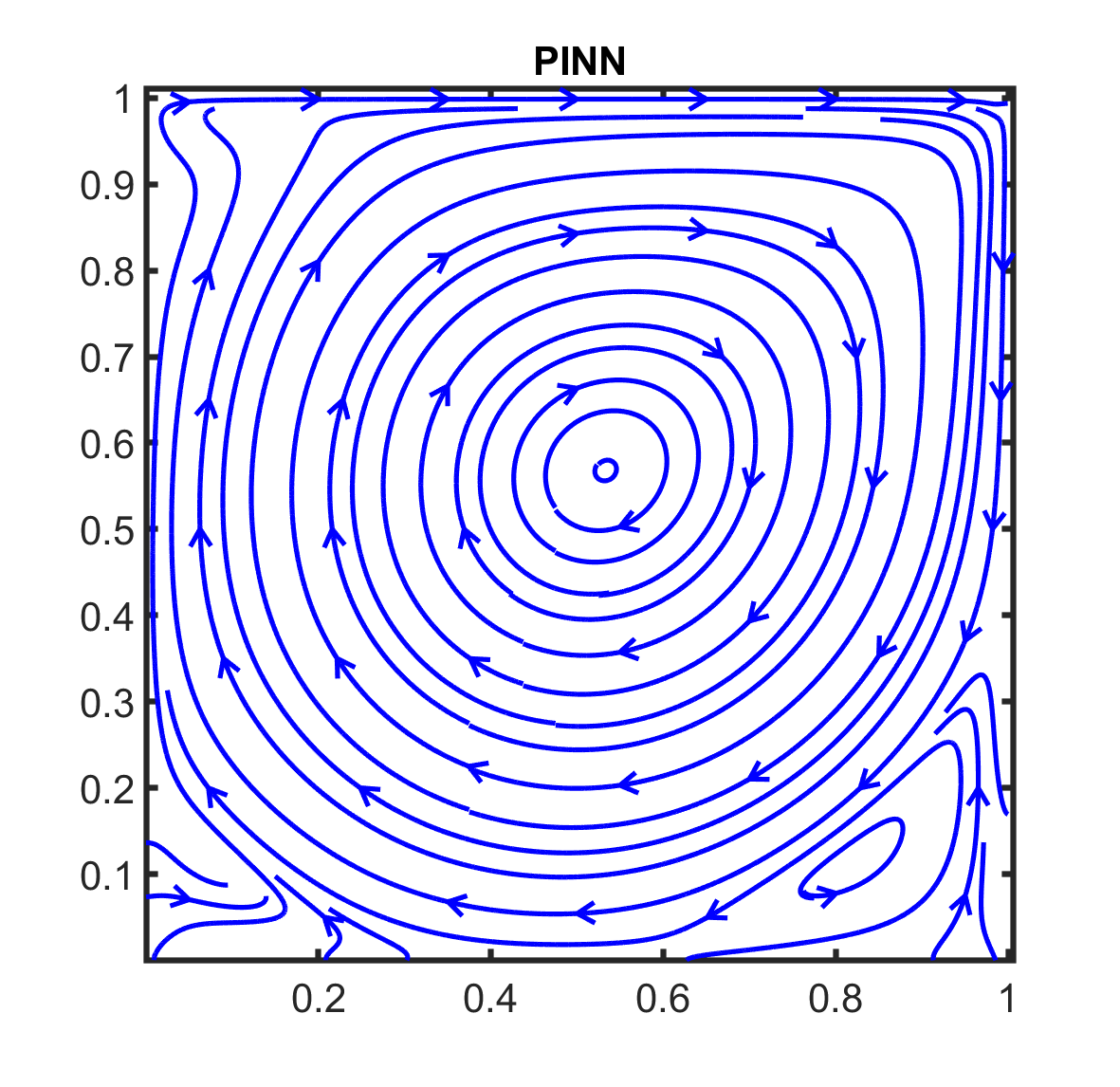}
\includegraphics[width=0.32\textwidth,trim=0 20 0 5,clip]{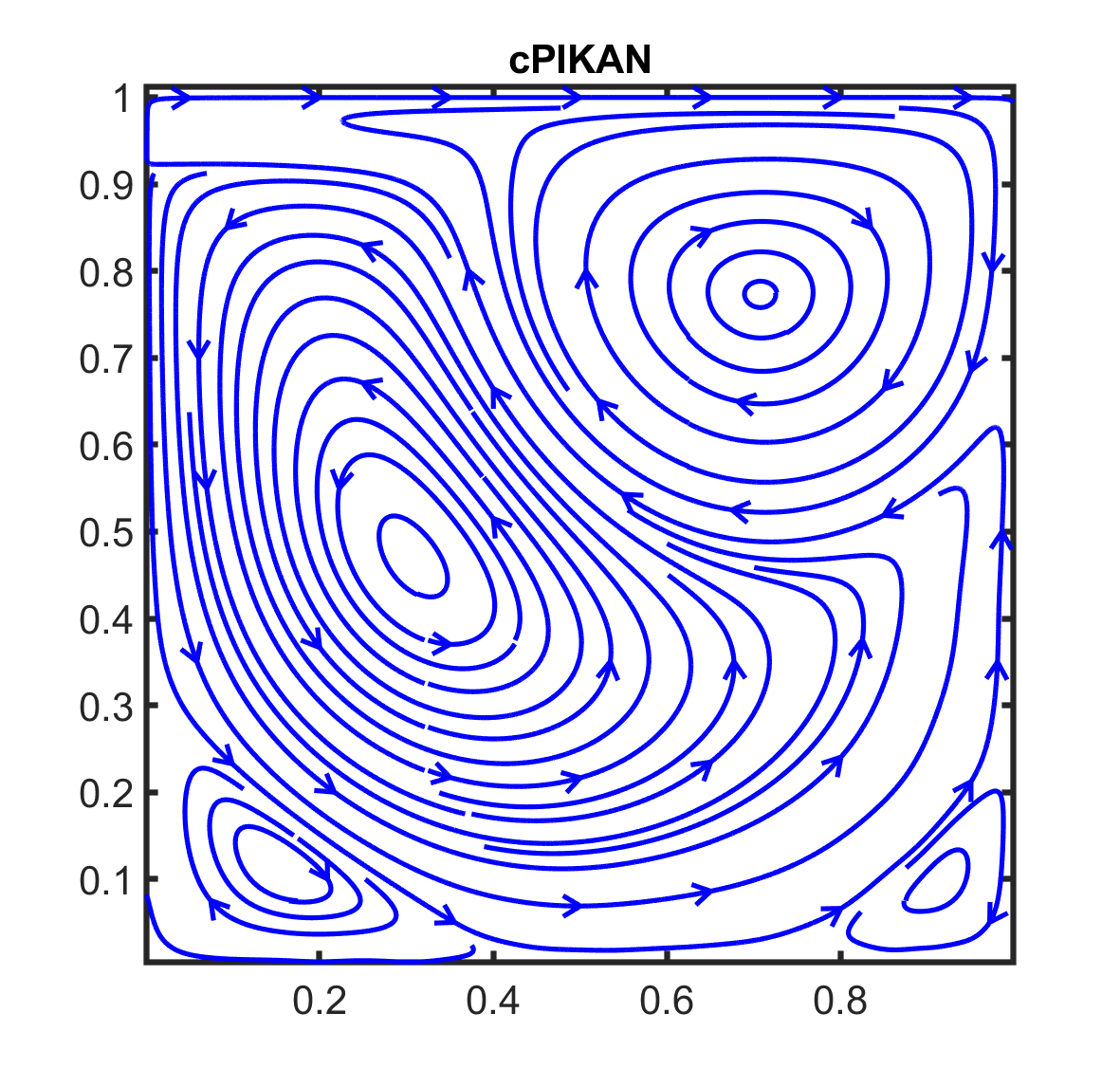}
\includegraphics[width=0.32\textwidth,trim=0 20 0 5,clip]{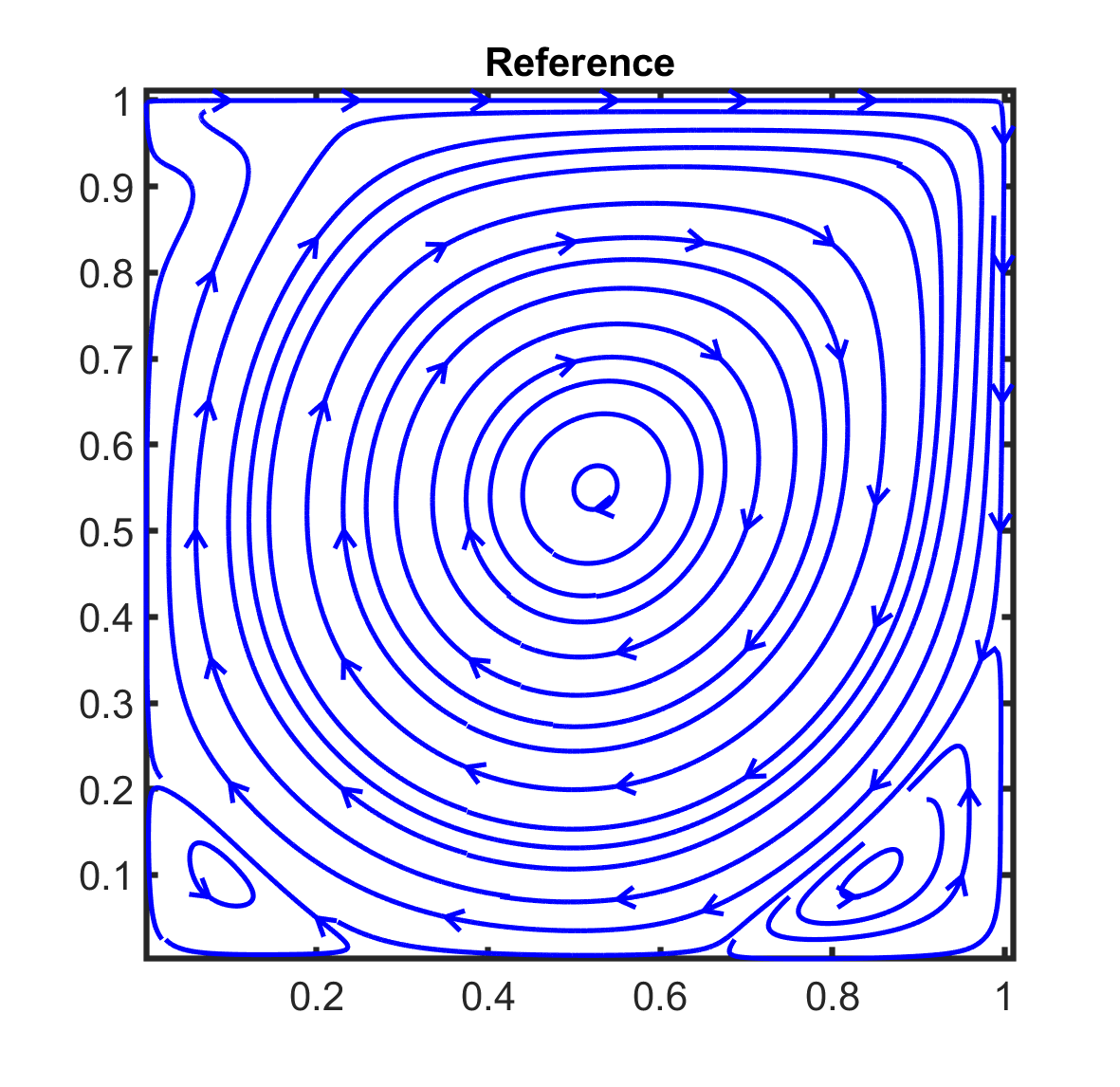}
\caption*{$Re=2000$.}
\caption{Streamlines of the cavity flow, inferred by the vanila PINN, cPIKAN, as well as the reference solution. }

\label{fig:streamlines_cavity}
\end{figure}

The inferred streamlines from the trained PINN/cPIKAN at final training stage are shown in \autoref{fig:streamlines_cavity}. It could be seen that at $Re=400$, both PINN and cPIKAN successfully reproduce the small eddies at left bottom and right bottom corners. The result of cPIKAN is sightly better than the counterpart of PINN, since the streamlines on the right bottom corners are in closed circles, while streamlines generated by PINN penetrate into the wall. At $Re=2000$, the streamlines generated by the cPIKAN are totally different from the reference solution, which indicates that the cPIKAN might stuck at a local minimum from the very beginning of the training. 

In summary, we found that PIKANs based on the Jacobi type of polynomials can generate comparable accurate solution to that of PINN, while the cost of training can be several times more. PIKANs can suffer from unstable training for the flow at high $Re$, but with the help of EVM or RBA, PIKANs can return to the correct training trajectory and generate accurate solution.

\subsection{Allen-Cahn equation}
In this example, we investigate the efficacy of PIKAN, cPIKAN, cPIKAN with RBA, and PINN with RBA for solving the 2D (1 + 1D) nonlinear Allen-Cahn equation \cite{allen1972ground}. The Allen-Cahn equation is expressed as follows,
\begin{align}\label{AC_Equation}
\frac{\partial u}{\partial t} -D \frac{\partial^2 u}{\partial x^2} + 5\left(u^3-u\right)=0,
\end{align}
where $D=1 \times 10^{-4}$ and $t \in[0,1],~x \in[-1,1]$ and with following initial and boundary conditions 
\begin{align*}
u(x, 0) &=x^2 \cos (\pi x) \\
u(-1, t) &=u(1, t)=-1.
\end{align*}

\begin{figure}
\centering
\includegraphics[trim={0cm 0cm 0cm 0cm}, clip, width=\columnwidth,keepaspectratio]{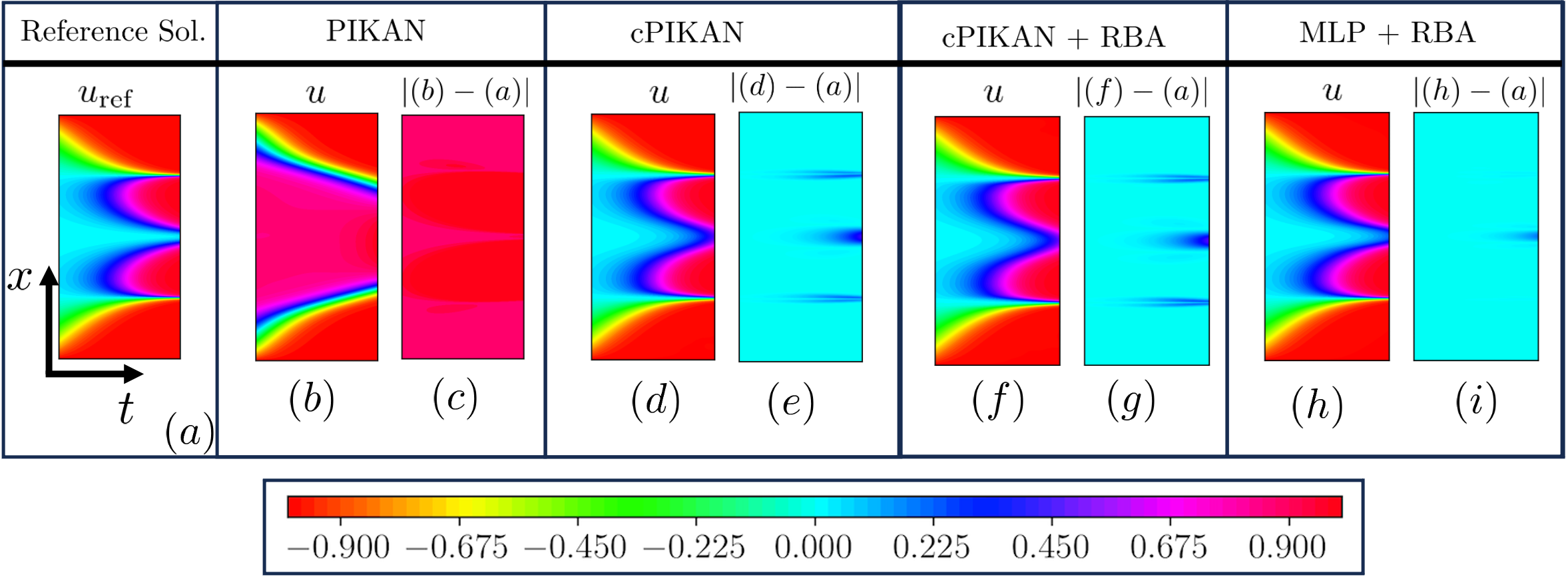}
\caption{The solution of the Allen-Cahn equation \eqref{AC_Equation} in the spatio-temporal domain of $(x \in [-1, 1],~t \in [0, 1])$ is depicted, with the reference solution shown in subfigure (a). The solutions obtained from (b) PIKAN, (d) cPIKAN, (f) cPIKAN+RBA, and (h) MLP architectures are also displayed. Subfigures (c), (e), (g), and (i) represent the absolute pointwise error between the reference solution and the solutions obtained from PIKAN, cPIKAN, cPIKAN with RBA, and MLP with RBA, respectively.}
\label{fig:AC_sol_KANs}
\end{figure}

The solution of \eqref{AC_Equation} using PIKAN, cPIKAN, cPIKAN with RBA, and PINN with RBA is obtained by minimizing the following loss function
\begin{align}\label{ac_loss}
\mathcal{L} = w_{ic} \mathcal{L}_{ic} + w_{b c} \mathcal{L}_{b c}+w_{p d e} \mathcal{L}_{p d e},
\end{align}
where $w_{ic}, ~w_{b c}$ and $w_{\text {pde }}$ are weights that balances the contribution of the averaged loss terms for initial conditions $\left(\mathcal{L}_{\text {ic }}\right)$, boundary conditions $\left(\mathcal{L}_{\text {bc }}\right)$ and PDE residuals $\left(\mathcal{L}_{\text {pde }}\right)$ which are described as follows, 

\begin{align}\label{loss_ac}
\begin{aligned}
\mathcal{L}_{ic} & =\left(\sum_{l=1}^{N_i}\left|\mathcal{R}_{l, ic}\right|\right)^2\\
\mathcal{L}_{bc} & = \left(\sum_{b=1}^2 \sum_{i=1}^{N_b}\left|\mathcal{R}_{i, b}\right|\right)^2, \\
\mathcal{L}_{\text {pde }} & =\left\langle\left(\alpha_j \cdot\left|\mathcal{R}_j\right|\right)^2\right\rangle_j,
\end{aligned}
\end{align}
where, $\langle\cdot\rangle$ is the mean operator, $\mathcal{R}_{i, b}$, $\mathcal{R}_{l, ic}$ and $\mathcal{R}_j$ are the residuals for boundary conditions, initial conditions and PDE at points $l, ~i$ and $j$, respectively. $\alpha_j$ are the RBA weights.

\begin{figure}
 \centering
 \includegraphics[width=0.8\textwidth]{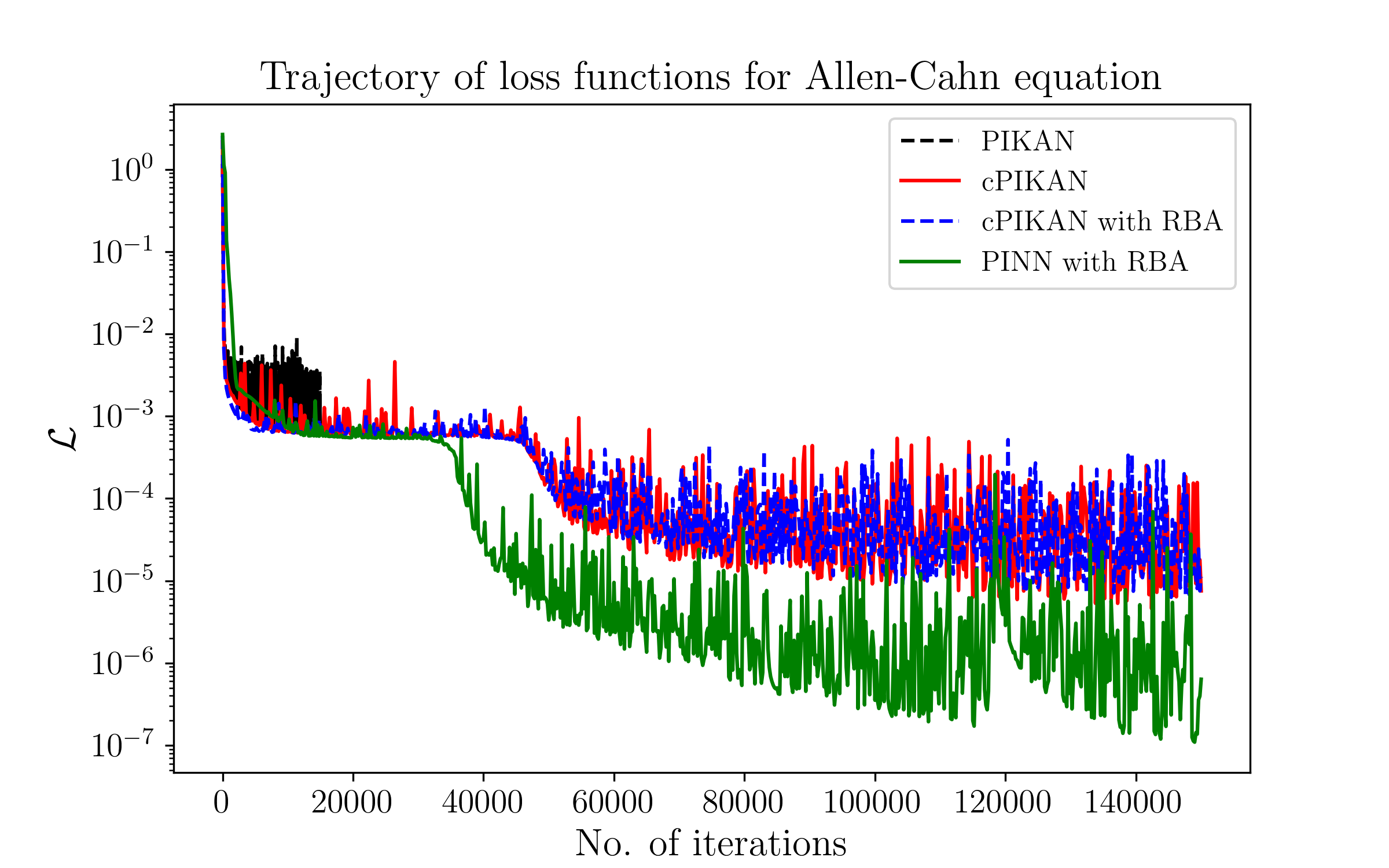}
 \caption{Loss functions showing the convergence of PIKAN, cPIKAN, cPIKAN with RBA and PINN with RBA while computing the solution of Allen-Cahn equation \eqref{AC_Equation}. It is to be noted that convergence PINN (MLP architecture) with RBA happened to be faster than any other method.}
 \label{fig:loss_AC_eqn}
\end{figure}

The solutions of \eqref{AC_Equation} using PIKAN, cPIKAN, cPIKAN with RBA, and PINN with RBA are shown in \autoref{fig:AC_sol_KANs}. The parameters used for training these networks are detailed in \autoref{tab:AC}. The results in \autoref{fig:AC_sol_KANs} were obtained using the Adam optimizer with a learning rate of \(5 \times 10^{-4}\). The training was performed as a single batch training till 150,000 iterations. In \autoref{fig:AC_sol_KANs}(a), we display the reference solution of \eqref{AC_Equation} computed using the spectral element method \cite{karniadakis2005spectral}. \autoref{fig:AC_sol_KANs}(b) presents the solution obtained using PIKAN, while \autoref{fig:AC_sol_KANs}(c) shows the absolute pointwise error between the reference and PIKAN solutions. 
The solution obtained from the PIKAN method did not converge to the reference solution, as  relative $l_2-$ error between PIKAN and reference solutions is 58.39\%. In \autoref{fig:AC_sol_KANs}(d) and (f), we show the solutions of \eqref{AC_Equation} obtained from cPIKAN and cPIKAN enhanced with RBA, respectively. The absolute pointwise errors in the solutions obtained from cPIKAN and cPIKAN with RBA are shown in \autoref{fig:AC_sol_KANs}(e) and (g), respectively. The relative $l_2-$ errors between the reference solution and those from cPIKAN and cPIKAN with RBA are 5.15\% and 5.65\%, respectively, indicating almost similar level of accuracy among them. In \autoref{fig:AC_sol_KANs}(h) and (i), we show the solution of \eqref{AC_Equation} obtained using PINN (MLP architecture) with RBA and the absolute pointwise error, respectively. 
The relative $l_2-$ error between the PINN and the reference solutions is 1.51\%. In \autoref{fig:loss_AC_eqn}, we show the convergence of all the methods by plotting the loss function \autoref{loss_ac} against the iterations. 
It is evident from \autoref{fig:loss_AC_eqn} that the MLP-based architecture, enhanced with RBA, exhibits faster convergence compared to the other methods. A detailed description of parameters, errors, and efficiency (in terms of runtime) is provided in \autoref{tab:AC}. The runtime measurements in \autoref{tab:AC} were taken on an Nvidia's GeForce RTX-3090 GPU. 
In \autoref{tab:AC} it is noted that runtime for cPIKAN and cPIKAN with RBA (Row 2 and 3) is almost similar despite having 50000 additional RBA parameters. This is caused by latency while moving the data from the DRAM to the processor. As the model and data are very small, the volatile GPU utility does not exceed more than 15\%. Therefore, the runtime for cPIKAN and cPIKAN with RBA is dominated by latency. 
\begin{table}[h]
\centering
\scalebox{0.90}{
\begin{tabular}{|c|c|c|c|c|}
\hline
Method & $[N_i, N_b, N_f]$ & No. of parameters & Relative $L^{2}$ error &Time: (ms/it)\\
\hline
PINN with RBA & [200, 100, 50000]  & 50,049 & 1.51 \%  & 22.93 ms\\
\hline
cPIKAN & [200, 100, 50000] & 6,720 & 5.15 \% & 39.31 ms \\
\hline
cPIKAN with RBA &  [200, 100, 50000]  & 56,720$^*$ & 5.65 \%  & 39.28 ms \\
\hline
PIKAN & [200, 100, 50000] & 6721 & 58.39 \% & 2633.12 ms\\
\hline
\end{tabular}
}
\caption{Details of (hyper) parameters used for solving Allen-Cahn equation using PINN with RBA, cPIKAN, cPIKAN with RBA and PIKAN based architecture. Here $N_i$, $N_b$ and $N_f$ represents the number of spatio-temporal points used for computing initial, boundary and residual value of $u$ during the training the networks. Time per iteration is measured on Nvidia's GeForce RTX-3090 GPU. $^*:$ 56720 parameters include 6720 model parameter and 50,000 trainable RBA weights.}
\label{tab:AC}
\end{table}

\subsection{Reaction-diffusion equation}

\begin{figure}[h!]
    \centering
    \subfigure[B-cPIKAN.]{
    \includegraphics[scale=.25]{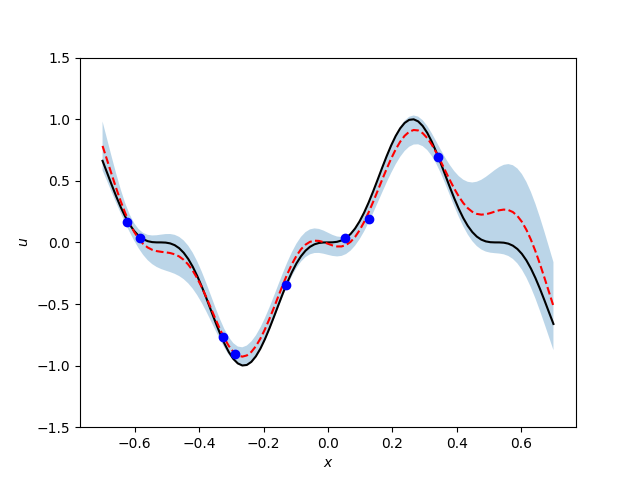}
    \includegraphics[scale=.25]{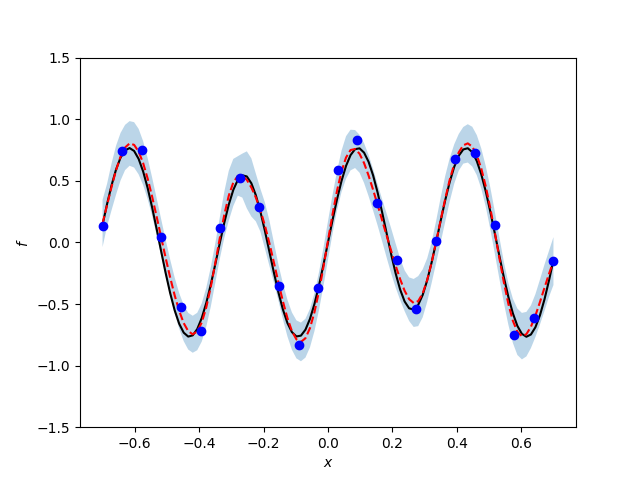}
    \includegraphics[scale=.25]{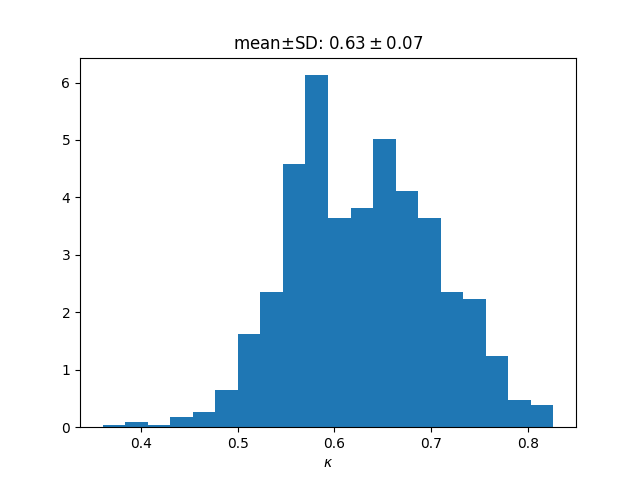}
    }
    \subfigure[B-PINN.]{
    \includegraphics[scale=.25]{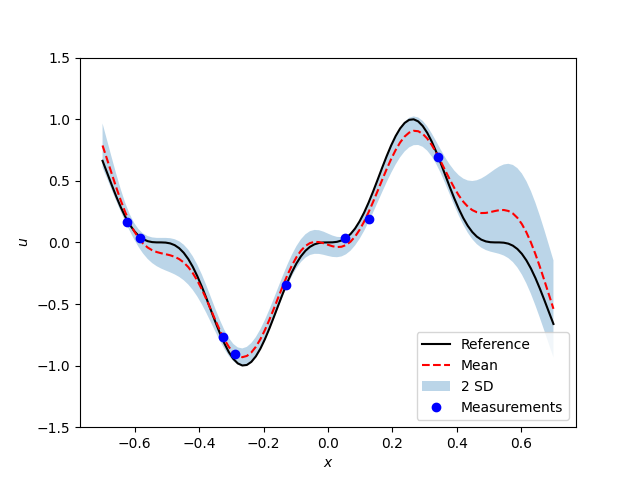}
    \includegraphics[scale=.25]{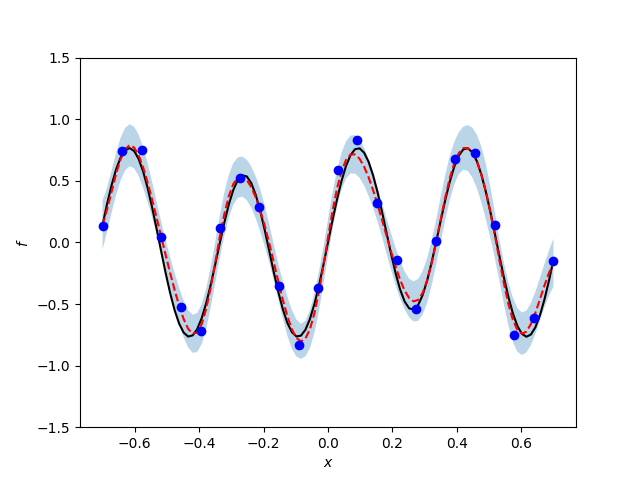}
    \includegraphics[scale=.25]{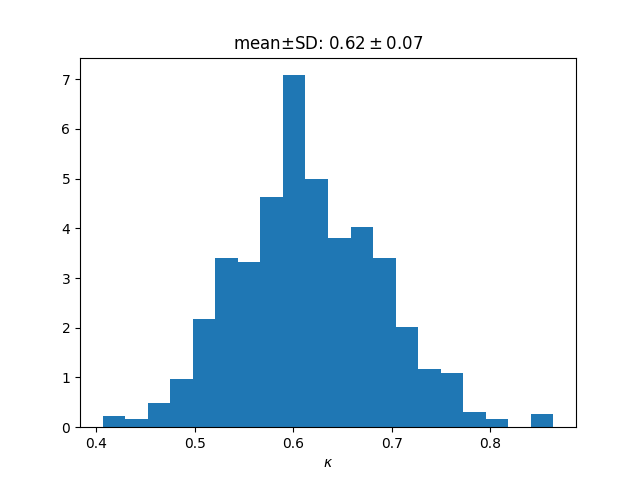}
    }
    \caption{Results from B-cPIKAN and B-PINN for the inverse problem on the 1D steady-state reaction-diffusion equation with noisy data of $u$ and $f$: from left to right are inferences of $u, f$, and $\kappa$. Here, the epistemic uncertainty of the network is quantified. The B-cPIKAN and B-PINN methods perform similarly in inferring $u$ and $f$. The B-cPIKAN method is able to provide slightly better inference over $\kappa$ (the exact is $0.7$): the error is lower and the uncertainty bounds the error.}
    \label{fig:bpinn}
\end{figure}

\begin{table}[h]
\centering
\begin{tabular}{|c|c|c|c|c|}
\hline
Methods & Rel. $l_2$--error of $u$ & Rel. absolute error of $\kappa$ & No. of parameters & Time \\
\hline
B-cPIKAN & $22.95\%$ & $10.04\%$ & 480 & 139s\\
\hline
B-PINN & $22.67\%$ & $11.90\%$ & 481 & 39s\\
\hline
\end{tabular}
\caption{Results from B-cPIKAN and B-PINN for the inverse problem on the 1D steady-state reaction-diffusion equation with noisy data of $u$ and $f$. The architecture of B-cPIKAN is $[1, 10, 10, 1]$ with degree three for the Chebyshev polynomials while the architecture of B-PINN is $[1, 20, 20, 1]$ with hyperbolic tangent activation function. HMC with the same hyperparameter is employed to sample from both posterior distributions on a standard laptop CPU (13th Gen Intel(R) Core(TM) i9-13900HX with 2.20 GHz processor).}
\label{tab:bpinn}
\end{table}

In this example, we extend PI-KAN to address noisy data in solving differential equations with uncertainty quantification (UQ) \cite{YANG2021109913, psaros2023uncertainty, zou2024neuraluq, meng2022learning, zou2023uncertainty, linka2022bayesian, yin2023generative, zou2024leveraging}. In particular, we equip cPIKAN with the Bayesian framework and estimate the posterior distribution of its parameters given data, i.e. $p(\theta|\mathcal{D})$, using Hamiltonian Monte Carlo (HMC) method \cite{neal2011mcmc} (see \cite{psaros2023uncertainty, zou2024neuraluq} for more details regarding UQ in SciML). We refer to the Bayesian cPIKAN as B-cPIKAN and demonstrate its capability by solving the following 1D steady reaction-diffusion equation with noisy data:
\begin{equation}\label{eq:bpinn}
    D \partial_{xx} u + \kappa \tanh(u) = f,
\end{equation}
where $f$ is the source term, $D=0.01$ denotes the diffusion rate, and $\kappa=0.7$ the reaction rate which is assumed to unknown.
The target is to infer $u$ and $\kappa$ with uncertainty given noisy data of $u$ and $f$, i.e. an inverse problem. 
Specifically, we choose $u(x) = \sin^3(6x), x\in[-0.7, 0.7]$ to be the exact and derive $f(x)$ analytically by plugging $u(x)=\sin^3(6x)$ into \eqref{eq:bpinn}. Eight measurements of $u$ are randomly sampled from $[-0.7, 0.7)$, following the uniform distribution, and corrupted by additive Gaussian noise with mean zero and standard deviation $0.05$, and 24 measurements of $f$ are uniformly sampled and corrupted by additive Gaussian noise with mean zero and standard deviation $0.1$.

We employ both the B-cPIKAN and B-PINN method to solve the inverse problem. For B-cPIKAN, the architecture of the network is $[1, 10, 10, 1]$ with degree three for the Chebyshev polynomials, while for B-PINN, the architecture is $[1, 20, 20, 1]$, such that the number of parameters of these two networks is approximately the same. The prior of the B-cPIKAN is chosen to be independent Gaussian with mean zero and standard deviation $0.5$, and the prior of the B-PINN is independent Gaussian with mean zero and standard deviation $1$.
We employ adaptive HMC with initial step size $0.01$ and $50$ leapfrog steps to sample from both posterior distributions. We set the number of burn-in samples to be $2,000$ and the number of posterior samples to be $1,000$. 
An open-source Python library NeuralUQ \cite{zou2024neuraluq} is utilized for fast and reliable implementation.
We note that here we only quantify the epistemic uncertainty of the network.
Results are presented in \autoref{fig:bpinn} and \autoref{tab:bpinn}, from which we can see the B-cPIKAN and B-PINN perform similarly: the predicted uncertainties are able to bound the errors between the predicted means and the exact. In particular, the predicted uncertainties of $u$ from both methods grow near $x=0.7$ due to lack of measurements. The B-cPIKAN method is able to provide slightly better inference over $\kappa$ with higher computational cost: the error is smaller and the uncertainty is able to bound the error.

\subsection{1D Burgers' equation}

\begin{table}[h]
\centering
\begin{tabular}{|c|c|c|c|c|}
\hline
Methods & Rel. $l_2$--error & No. of parameters & Time: ms/iter\\
\hline
DeepONet & $5.83\% \pm 0.19\%$ & 63900 & $1.9$ \\
\hline
DeepOKAN 1 & $2.71\% \pm 0.08\%$ & 252800 & $3.9$\\
\hline
DeepOKAN 2 & $3.02\%\pm 0.13\%$ & 76400 & $3.9$ \\
\hline
\end{tabular}
\caption{Rel. $l_2$--error for learning the solution operator of the 1D Burgers' equation with viscosity $\nu=\frac{1}{100\pi}$. The architectures of the branch and trunk nets are $[128, 100, 100, 100, 100]$ and $[4, 100, 100, 100]$, respectively, for DeepONet and DeepOKAN 1, and are $[128, 50, 50, 50, 50]$ and $[4, 50, 50, 50]$ for DeepOKAN 2. Here Chebyshev KAN \cite{chebykan} is employed with degree three for DeepOKANs and hyperbolic tangent is used as the activation function for the DeepONet. The time per iteration is measured on an Nvidia's GeForce RTX-3090 GPU with 24 GB of memory.
}
\label{tab:burgers}
\end{table}

\begin{table}[h]
\centering
\begin{tabular}{|c|c|c|c|}
\hline
Methods & $1\%$ noise & $5\%$ noise & $10\%$ noise \\
\hline
DeepONet & $5.83\%\pm 0.19\%$ & $5.93\%\pm0.18\%$ & $6.29\%\pm0.19\%$\\
\hline
DeepOKAN 1 & $2.72\%\pm0.08\%$ & $2.94\%\pm0.09\%$ & $3.57\%\pm0.05\%$\\
\hline
DeepOKAN 2 & $3.02\%\pm0.13\%$ & $3.24\%\pm0.12\%$ & $3.91\%\pm0.13\%$ \\
\hline
\end{tabular}
\caption{Rel. $l_2$--error for learning the solution operator of the 1D Burgers' equation when trained with clean data but tested with noisy input data. Here the noise is additive Gaussian noise with mean zero and different levels of standard deviation, and the noise level is defined as the percentage of the absolute value of the input function evaluated at the grid.}
\label{tab:burgers2}
\end{table}

We consider the 1D Burgers' equation with periodic boundary conditions:
\begin{equation}\label{burger_1D}
    \partial_tu + u\partial_x u = \nu \partial_{xx}u, x\in [0, 1], t\in[0, 1].
\end{equation}
where $\nu=\frac{0.01}{\pi}$ denotes the viscosity. In this example, we learn the surrogate operator for solution of equation \ref{burger_1D}, which maps an arbitrary initial condition sampled from a distribution, denoted as $u_0$, to the solution of equation \ref{burger_1D} at $t=1$:
\begin{equation}
    G: u_0(x) \mapsto u(x, t=1).\nonumber
\end{equation}

The training and testing data are generated following \cite{li2020fourier} where the initial condition is sampled from a Gaussian process defined as $\mu = N(0, 49^2(-\Delta+49I)^{-2.5})$ with embedded periodic boundary condition. A spatial resolution with $128$ uniform grids is used to resolve the input and output functions.
We apply four Fourier basis 
$$\{\cos(2\pi x), \sin(2\pi x), \cos(4\pi x), \sin(4\pi x)\}$$
to the input of the trunk net and data normalization to the output of DeepONets to improve the performnace \cite{lu2022comprehensive}.
$1,000$ and $200$ functions of $u_0$ and $u(\cdot, t=1)$ are used for training and testing, respectively, and we normalize the output of the operator network based on the mean and standard deviation of the training data.

We employ one DeepONet and two DeepOKANs to learn the solution operator $G$. Specifically, the architecture of the DeepONet is $[128, 100, 100, 100, 100]$ for the branch net and $[4, 100, 100, 100]$ for the trunk net, both of which are equipped with $\tanh$ activation function, while the architectures of DeepOKANs (DeepOKAN 1/2) are [128, 100, 100, 100, 100]/[128, 50, 50, 50, 50] for the branch net and [4, 100, 100, 100]/[4, 50, 50, 50] for the trunk net. Both DeepOKANs are based on the Chebyshev KAN \cite{chebykan} and have degree three for the Chebyshev polynomials. The Adam optimizer \cite{kingma2014adam} is employed for all operator networks. The learning rate for the DeepONet is 1e-3 for 100k iterations and 1e-4 for another 100k iterations, while for both DeepOKANs it is 1e-4 for 100k iterations and 1e-5 for another 100k iterations. 
To avoid overfitting, we apply a $l_2$ regularizer \cite{loshchilov2017decoupled} with weighting coefficient 1e-5 to both operator networks.
The errors are presented in \autoref{tab:burgers}. We observe that DeepOKANs perform significantly better than the DeepONet at higher computational cost.

We further test the robustness of these operator networks against noisy input functions. Specifically, networks are trained with clean data while tested with noisy data. Here we consider Gaussian noise with mean zero added to the value of $u_0(x_i), i=1,...,N_v$ where $x_i$ are the uniform grids on which the input and output functions are resolved. The standard deviation of the noise is proportional to the absolute value of $u_0(x_i)$. We test three noise levels with $1\%$, $5\%$ and $10\%$ and present results in \autoref{tab:burgers2}. We observe that DeepOKANs are more robust to noisy input functions compared to the DeepONet.

\subsection{120-dimensional Darcy problem}

\begin{table}[h]
\centering
\begin{tabular}{|c|c|c|c|}
\hline
Methods & $L^2$ relative error & No. of parameters & Time: ms/iter\\
\hline
DeepONet & $1.62\%\pm 0.15\%$ & 147000 & $2.3$ \\
\hline 
DeepOKAN & $2.18\%\pm0.02\%$ & 585200 & $8.8$ \\
\hline
\end{tabular}
\caption{Rel. $l_2$--error for learning the solution operator of the Darcy problem. The architectures of the branch and trunk nets are $[961, 100, 100, 100, 100]$ and $[2, 100, 100, 100]$, respectively, for both operator networks. Chebyshev KAN \cite{chebykan} is employed with degree three for the DeepOKAN and hyperbolic tangent is used as the activation function for the DeepONet. The time per iteration is measured on an Nvidia's GeForce RTX-3090 GPU with 24 GB of memory.}
\label{tab:darcy}
\end{table}

\begin{table}[h]
\centering
\begin{tabular}{|c|c|c|c|}
\hline
Methods & $1\%$ noise & $5\%$ noise & $10\%$ noise \\
\hline
DeepONet & $1.66\%\pm 0.15\%$ & $2.25\%\pm 0.12\%$ & $3.47\%\pm0.12\%$\\
\hline 
DeepOKAN & $2.18\%\pm0.02\%$ & $2.20\%\pm 0.02\%$ & $2.30\%\pm 0.03\%$\\
\hline
\end{tabular}
\caption{Rel. $l_2$--error for learning the solution operator of the Darcy problem when trained with clean data but tested with noisy input data. The noise is additive Gaussian noise with mean zero and different levels of standard deviation, and the noise level is defined as the percentage of the absolute value of the input function evaluated at the grid.}
\label{tab:darcy2}
\end{table}

In this section, we consider a 2D steady-state flow through porous media described by steady-state Darcy's law expressed as
\begin{equation}
    \nabla\cdot(\lambda(x, y)\nabla u(x, y)) = f, x, y \in (0, 1),
\end{equation}
where the source term $f = -30$.
Here $\lambda$ denotes the hydraulic conductivity field and $u$ the hydraulic head. 
The boundary condition is specified as follows:
\begin{equation}
    u(0, y) = 1, u(1, y) = 0,
    \partial_{\mathbf{n}}u(x, 0) = \partial_{\mathbf{n}}u(x, 1) = 0.
\end{equation}
We learn the solution operator which maps $\log(\lambda)$ to $u$:
\begin{equation}
    G: \log(\lambda)(x, y) \mapsto u(x, y),\nonumber
\end{equation}
using dataset from \cite{psaros2023uncertainty, zou2024neuraluq, zou2023uncertainty}, in which the logarithm of the conductivity is sampled from a truncated Karhunen-Lo\`eve expansion of a Gaussian process with zero mean and the following kernel:
\begin{equation}
    k(x, x^\prime, y, y^\prime) = \exp(-\frac{(x-x^\prime)^2}{2l^2} - \frac{(y - y^\prime)^2}{2l^2}), x, x^\prime, y, y^\prime \in [0, 1],\nonumber
\end{equation}
where $l=0.25$ denotes the correlation lengths. We use a $31\times31$ uniform grid to represent $\log(\lambda)$ and $u$, and the first $120$ leading terms of the expansion is kept. To perform training and testing, 
$10,000$ and $1,000$ functions of $\log(\lambda)$ and $u$ are used, respectively. We normalize both the input and output of the operator network based on the mean and standard deviation of the training data for better performance.

We employ one DeepONet and one DeepOKAN to learn the solution operator $G$. The architectures are $[961, 100, 100, 100, 100]$ for the branch net and $[2, 100, 100, 100]$ for the trunk net, for both DeepONet and DeepOKAN. The DeepONet has $\tanh$ as the activation function, while the DeepOKAN is based on Chebyshev KANs and has degree three for the Chebyshev polynomials.
The Adam optimizer \cite{kingma2014adam} is employed for both operator networks. The learning rate for the DeepONet is 1e-3 for 100k iterations and 1e-4 for another 100k iterations, while for the DeepOKAN it is 1e-4 for 100k iterations and 1e-5 for another 100k iterations. We impose a $l_2$ regularizer \cite{loshchilov2017decoupled} with weighting coefficient 1e-4 to the DeepOKAN to avoid the overfitting.
The errors are presented in \autoref{tab:darcy}, from which we observe that the DeepONet outperforms the DeepOKAN at lower computational cost. However, as shown in \autoref{tab:darcy2}, the robustness of the DeepOKAN is still better than the DeepONet when they are trained with clean input data while tested with noisy input data. In particular, the DeepOKAN becomes better in accuracy as the noise level grows larger. 





%% file: sec_4.tex
\section{Learning in PIKANs}
\label{Sec6}

\subsection{Information bottleneck method}
The Information Bottleneck (IB) method offers a perspective on the training and performance of neural networks using principles from information theory. It lays out a framework for determining the ideal balance between compression and prediction in supervised learning, proposing a principle for forming a condensed representation of layer activations $\mathcal{T}$ with respect to an input variable $\mathcal{X}$, which preserves as much information as possible about an output variable $\mathcal{Y}$ \cite{tishby2000information,tishby2015deep}. Central to this theory is the use of mutual information $I(x,y)$, a measure of the information one random variable $(y)$ reveals about another $(x)$. This indicates that the best model representations should retain all relevant information about the output while omitting irrelevant information from the inputs, thereby establishing an ``information bottleneck." A notable insight from IB is that deep learning progresses through two distinct phases: fitting and diffusion, delineated by a phase transition driven by the signal-to-noise ratio (SNR) of the gradients \cite{shwartz2017opening,goldfeld2020information,shwartz2022information}. The theory posits that significant learning happens during the gradual diffusive phase, crucial for the model's ability to generalize effectively. Anagnostopoulos et al.\cite{anagnostopoulos2024learning} extended the information bottleneck method to interpret how physics-informed neural networks learn and prosed the existence of a third phase denominated total diffusion. In this section, we will apply this framework to describe the learning dynamics of cPIKANs and PINNs.

\subsection{Signal-to-noise ratio(SNR)}

As described in \cite{anagnostopoulos2024learning,shwartz2017opening,goldfeld2020information,shwartz2022information}, the batch-wise signal-to-noise ratio (SNR) is a metric used to identify the training dynamics of neural networks and can be described as follows:

\begin{equation} \text{SNR} = \frac{\lVert \mu \rVert_{2}}{\lVert \sigma \rVert_{2}} = \frac { \lVert \mathbb{E}[\nabla_{\theta}{\mathcal{L}_{\mathcal{B}}}] \rVert_{2}} {\lVert std[\nabla_{\theta}{\mathcal{L}_{\mathcal{B}}}] \rVert_{2}}
\label{SNR}
\end{equation}

\noindent where $\theta$ are the network parameters, and $\lVert \mu \rVert_{2}$ and $\lVert \sigma \rVert_{2}$ are the $L^2$ norms of the batch-wise mean and standard deviation of the total loss gradients ($\nabla_{\theta}{\mathcal{L}_{\mathcal{B}}}$). Under this definition, the ``signal" represents an idealized gradient that drives the optimizer to minimize the error of all subdomains, and the noise is the perturbation from the ideal gradient related to learning from the average of a finite number of observations. Following \cite{anagnostopoulos2024learning}, we analyze the PINN and cPIKAN training dynamic of full-batch Adam trained with the entire dataset $\mathcal{X}$, and calculate $\nabla_{\theta}{\mathcal{L}_{\mathcal{B}}}$ without performing an update of $\theta$ for each i.i.d.  batch ($\mathcal{B}$) so that we can investigate the batch-wise behavior on the same iteration $t$. 

\subsection{Stages of Learning}

The different stages can be interpreted as a process where the model fits the data (captures relevant information) and then compresses it (discards irrelevant information), further enhancing its generalization ability \cite{anagnostopoulos2023residual}. Each phase is characterized by the dominant term in the SNR. Highly deterministic regimes are characterized by a high signal and, with it, a high SNR. On the other hand, highly stochastic stages are defined by high noise and low SNR.

\begin{figure}[H]
 \centering
\includegraphics[width=\textwidth]{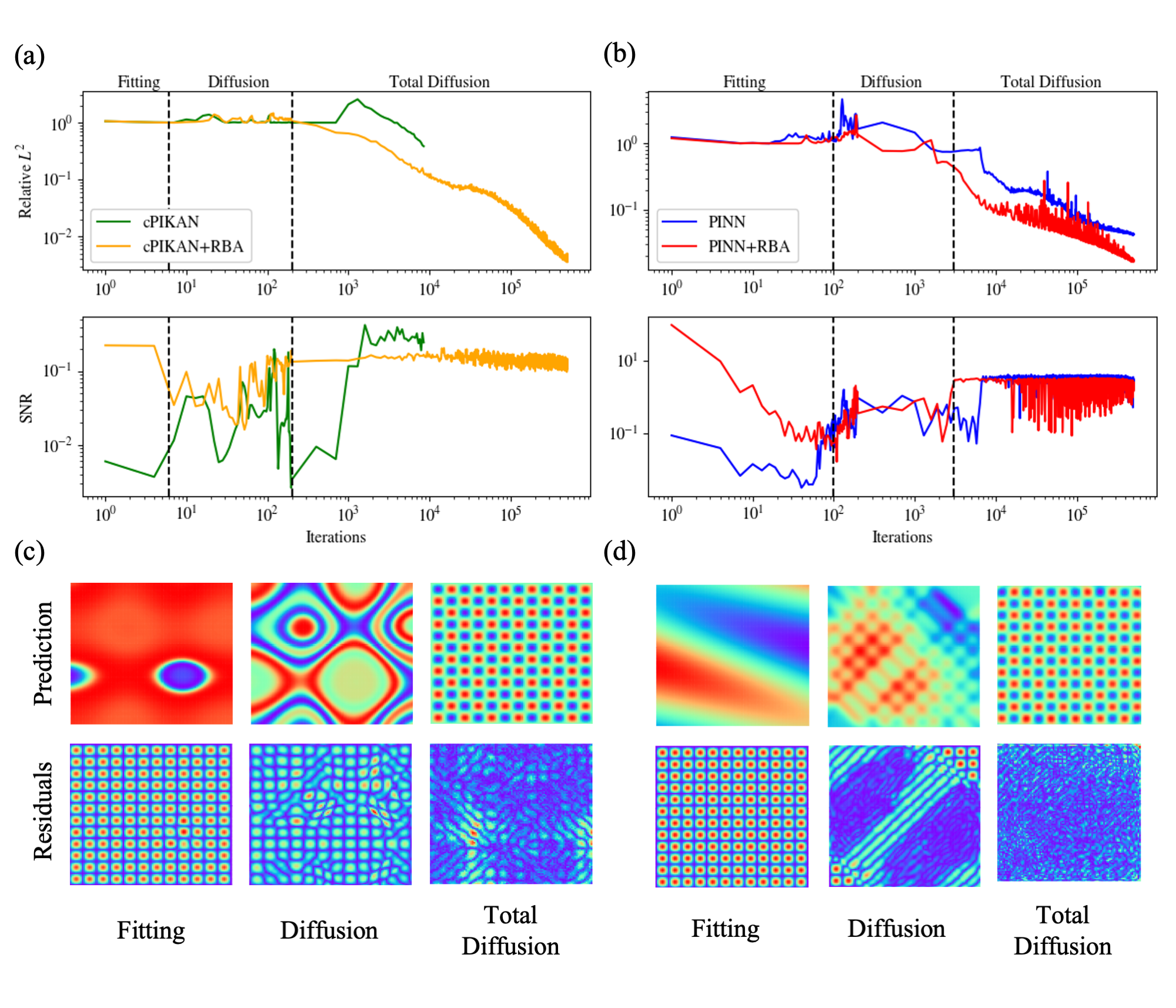}
 \caption{Training dynamics and stages of learning.  Relative $L^2$ convergence and corresponding signal-to-noise ratio (SNR) for (a) cPIKANs and (b) PINNs. The vertical dashed lines indicate the transitions of cPIKANs+RBA and PINNs+RBA. During \textbf{fitting}, PIKANs and PINNs SNR go from high to low. This suggests an initial phase where the model closely fits the training data. The \textbf{Diffusion} phase is considered an exploratory stage characterized by a fluctuating low SNR. In the last stage, \textbf{total diffusion}, the SNR suddenly increases and converges to a critical value, and the generalization (i.e., relative  $L^2$) error decreases faster, suggesting an optimal convergence. Notice that the best-performing models transition to total diffusion faster. Even though cPIKAN became undefined during the initial iterations, the three stages of learning are still identifiable.  Prediction at residual distributions at different stages of learning in (c) cPIKAN and (d) PINN. The \textbf{fitting} phase is highly deterministic, so the residuals display an ordered pattern. As the SNR decreases and the model transitions to a stochastic \textbf{diffusion}, the residuals gradually become disordered. Finally, in \textbf{total diffusion} the model
 reaches an equilibrium state, simplifies internal representations, and reduces their complexity, making the model much more efficient and generalizable. Notice that during this stage, the predictions closely match the analytical solution. This phase is characterized by highly stochastic (i.e., noisy) residuals. } 
 \label{fig:SNR}
\end{figure}

\paragraph{Fitting}  At the beginning of training, the loss and its gradients are high for all subdomains. This agreement induces an initial high SNR  characterized by a signal (i.e., direction) that helps the model reduce the training error of all subdomains. However, as the loss and its gradients (i.e., signal) decrease, the disagreement between subdomains  (i.e., noise) increases too, which induces a low SNR. Therefore, the fitting stage can be defined as a deterministic phase characterized by a transition from high to low SNR. As shown in Figure~\ref{fig:SNR}(a) and (b), cPIKANs and PINNs present a fitting stage. In particular, their SNR goes from high to low, and their corresponding residuals display an ordered pattern (See Figure~\ref{fig:SNR}(c) and (d)).

\paragraph{Diffusion} Once the model has learned to fit the data (i.e., general traits), it starts an exploration stage aiming to find a signal (i.e., direction) that minimizes the training error in all subdomains. During this stage, the network weights start diffusing and aim to improve the model's generalization capabilities, breaking the initial states' order. Thus, the diffusion stage is characterized by a low fluctuating SNR. Figure~\ref{fig:SNR}(a) and (b) show that cPIKANs and PINNs display a diffusion state. Notice that in this stochastic stage, the residuals become disordered (i.e., Figure~\ref{fig:SNR}(a) and (b)), and SNR starts to oscillate.

\paragraph{Total Diffusion} Once the model has identified an optimal signal, the SNR suddenly increases to an equilibrium stage where the model exploits a consistent direction that minimizes the generalization error in all subdomains. During this phase, the model simplifies the internal representations of the learned patterns by keeping the important features and discarding the irrelevant ones, thus effectively reducing its complexity and breaking the order of the corresponding residuals. As shown in Figure~\ref{fig:SNR}(a) and (b), PINNs and PIKANs display a total diffusion stage. Notice that, for all representation models, as soon as the optimal direction (i.e., signal) is found (i.e., total diffusion starts), the generalization error (i.e., relative $L^2$) decreases faster, indicating optimal convergence. Thus, unsurprisingly, the best-performing models (i.e., PINN+RBA and cPIKAN+RBA) transition to total diffusion first.

